\documentclass[10pt,twocolumn,letterpaper]{article}

\usepackage[pagenumbers]{cvpr} %

\definecolor{darkorange}{rgb}{1.0, 0.55, 0.0}

\newcommand{\cellfirst}[1]{\textbf{#1}}
\newcommand{\cellsecond}[1]{\underline{#1}}
\newcommand{\cellthird}[1]{{#1}}

\newcommand{\supp}{\emph{Appendix}}

\newcommand{\method}{Elastic Latent Interface Transformer}
\newcommand{\methodacronym}{ELIT}

\newcommand{\ourmodel}{}

\newcommand{\generator}{\mathcal{G}}

\newcommand{\readlayer}{\mathcal{R}}
\newcommand{\writelayer}{\mathcal{W}}

\newcommand{\inputtensor}{\mathbf{X}}
\newcommand{\inputtensortime}{\mathbf{X}_{t}}
\newcommand{\inputtensornoise}{\mathbf{X}_{0}}
\newcommand{\inputtensorclean}{\mathbf{X}_{1}}
\newcommand{\conditioning}{{c}}

\newcommand{\velocity}{{v}}
\newcommand{\velocitypred}{{\hat{v}}}
\newcommand{\spatialact}{{s}}
\newcommand{\latentact}{{l}}

\newcommand{\latentactca}{{l}_{CA}}

\newcommand{\spatialactout}{{s}_{O}}
\newcommand{\latentactout}{{l}_{O}}

\newcommand{\hidsize}{{d}}
\newcommand{\numblocks}{{B}}
\newcommand{\numblocksstart}{{B}_\mathrm{in}}
\newcommand{\numblocksmid}{{B}_\mathrm{core}}
\newcommand{\numblocksend}{{B}_\mathrm{out}}

\newcommand{\numgroups}{{G}}
\newcommand{\latenttokcount}{{K}}
\newcommand{\latenttokcountgroup}{{J}}
\newcommand{\latenttokcountgroupmasked}{{\tilde{J}}}
\newcommand{\latenttokcountgroupmin}{{J_\mathrm{min}}}
\newcommand{\latenttokcountgroupmax}{{J_\mathrm{max}}}

\newcommand{\latenttokcountgroupweak}{\tilde{J}_{\mathrm{w}}}

\newcommand{\numtokens}{N}

\newcommand{\tradeoffeff}{\rho}

\newcommand{\datadistribution}{p_{d}}
\newcommand{\timestepdistribution}{p_{t}}
\newcommand{\noisedistribution}{p_{n}}
\newcommand{\difftimestep}{t}

\newcommand{\fidtenk}{\text{FID}_{10\text{K}}}
\newcommand{\fddtenk}{\text{FDD}_{10\text{K}}}
\newcommand{\fvdtenk}{\text{FVD}_{10\text{K}}}
\newcommand{\fidfifk}{\text{FID}_{50\text{K}}}
\newcommand{\fddfifk}{\text{FDD}_{50\text{K}}}

\newcommand\nnfootnote[2]{%
  \begin{NoHyper}
  \renewcommand\thefootnote{#1} %
  \footnotetext{#2}%
  \renewcommand\thefootnote{\arabic{footnote}} %
  \addtocounter{footnote}{-1}%
  \end{NoHyper}
}

\newcommand{\update}[1]{{\color{blue}#1}}
\renewcommand{\update}[1]{#1}

\definecolor{cvprblue}{rgb}{0.21,0.49,0.74}
\usepackage[pagebackref,breaklinks,colorlinks,allcolors=cvprblue]{hyperref}

\usepackage{booktabs,multirow,makecell,array}
\usepackage[table]{xcolor}
\usepackage{colortbl}
\usepackage{adjustbox}
\usepackage{amsmath}
\usepackage{hyperref}
\usepackage{url}
\usepackage{booktabs}
\usepackage{multirow}
\usepackage{amssymb,graphicx}
\usepackage{amssymb,pifont}
\usepackage{pgfplots}
\usepackage{caption}
\usepackage{subcaption}
\usepackage[table]{xcolor}
\usepackage{enumitem}
\usepackage{wrapfig}   %
\usepackage{booktabs}      %
\usepackage{siunitx}       %
\usepackage{multirow}      %
\usepackage{xcolor}        %
\usepackage{colortbl}      %
\usepackage{amsmath}
\usepackage{array}         %

\usepackage{booktabs}
\usepackage{tabularx}
\usepackage{wrapfig}
\usepackage{etoc}

\def\paperID{XXXXXX} %
\def\confName{CVPR}
\def\confYear{2026}

\title{One Model, Many Budgets: Elastic Latent Interfaces for Diffusion Transformers}

\author{%
  \makebox[\textwidth][c]{%
    \text{Moayed Haji-Ali}$^{1,2,\dagger}$ \qquad
    \text{Willi Menapace}$^{2}$ \qquad
    \text{Ivan Skorokhodov}$^{2}$ \qquad
    \text{Dogyun Park}$^{2, \dagger}$ \qquad
    \text{Anil Kag}$^{2}$ \qquad
  }\\[0.3em]
  \makebox[\textwidth][c]{%
    \text{Michael Vasilkovsky}$^{2}$ \qquad
    \text{Sergey Tulyakov}$^{2}$ \qquad
    \text{Vicente Ordonez}$^{1}$ \qquad
    \text{Aliaksandr Siarohin}$^{2}$%
  }\\[1em]
  \centering
  \parbox{\linewidth}{\centering
    ${}^1\text{Rice University}$ \qquad \qquad ${}^2\text{Snap Inc.}$\\
    \vspace{0.75em}
    Project Webpage: \href{https://snap-research.github.io/elit}{\color{blue}https://snap-research.github.io/elit}
  }
}

\begin{document}

\newcolumntype{C}[1]{>{\centering\arraybackslash}p{#1}}

\twocolumn[{
\renewcommand\twocolumn[1][]{#1}%
\maketitle
\vspace{-0.2in}
\begin{center}
\vspace{-12pt}
    \centering
    \includegraphics[width=\textwidth]{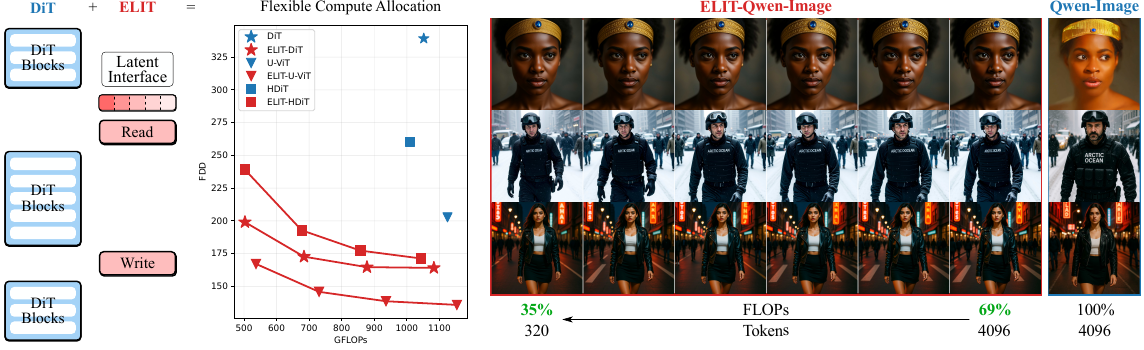}
   \vspace{-0.1in}
\vspace{-0.1in}
    \captionof{figure}{
    \textbf{Flexible compute allocation with \methodacronym{}.} Starting from a vanilla DiT, we add a variable-length set of latent tokens—the \emph{latent interface}—and two lightweight cross-attention layers, Read and Write. At inference, the number of latent tokens is a user-controlled knob that yields a smooth quality–FLOPs trade-off across DiT, U-ViT, HDiT, and MM-DiT backbones.
    }
    \label{fig:teaser}
\end{center}%
}]

\begin{abstract}
\nnfootnote{$\dagger$}{Work partially done during an internship at Snap Inc.}
Diffusion transformers (DiTs) achieve high generative quality but lock FLOPs to image resolution, limiting principled latency-quality trade-offs, and allocate computation uniformly across input spatial tokens, wasting resource allocation to unimportant regions. We introduce Elastic Latent Interface Transformer (ELIT), a drop-in, DiT-compatible mechanism that decouples input image size from compute. Our approach inserts a latent interface, a learnable variable-length token sequence on which standard transformer blocks can operate. Lightweight Read and Write cross-attention layers move information between spatial tokens and latents and prioritize important input regions. By training with random dropping of tail latents, \methodacronym{} learns to produce importance-ordered representations with earlier latents capturing global structure while later ones contain information to refine details. At inference, the number of latents can be dynamically adjusted to match compute constraints. \methodacronym{} is deliberately minimal, adding two cross-attention layers while leaving the rectified flow objective and the DiT stack unchanged. Across datasets and architectures (DiT, U-ViT, HDiT, MM-DiT), ELIT delivers consistent gains. On ImageNet-1K 512px, \methodacronym{} delivers an average gain of $35.3\%$ and $39.6\%$ in FID and FDD scores.

\vspace{-1em}

\end{abstract}

\section{Introduction}

Recent years have seen dramatic progress in image and video generation. Owing to the simplicity of their design, architectures centered on Diffusion Transformers (DiTs)~\citep{dit} have scaled reliably and delivered state-of-the-art fidelity~\citep{wu2025qwenimage,wan,avlink}. 
Compute has been the primary determinant of generation quality, but continued scaling has inflated training and inference costs. This raises a central question: do DiTs utilize available computation effectively? We argue that such costs are amplified by the \emph{rigidity} in the DiT design. First, a DiT typically commits to a per-step computational cost that is a fixed function of the input resolution, without accounting for latency and compute constraints. Second, we found that DiT allocates computation uniformly across image regions. In a controlled experiment, we probe the ability of a DiT to use extra compute to improve generation quality. As expected, quality improves on standard images when we increase the number of tokens by reducing the patchification size. However, when we increase the number of tokens by padding encoded image patches with zero-valued patches, we find that DiT fails to leverage the extra computation 
to improve generation quality. These observations suggest that compute is spent \emph{uniformly} across image tokens. This is suboptimal since visual information in images is uneven: some regions are easy, others require more work.  In this context, \emph{learning} how to allocate computation across tokens through a flexible architecture holds the potential to dynamically reduce unnecessary computation.

\begin{figure*}
    \centering
    \includegraphics[width=\linewidth]{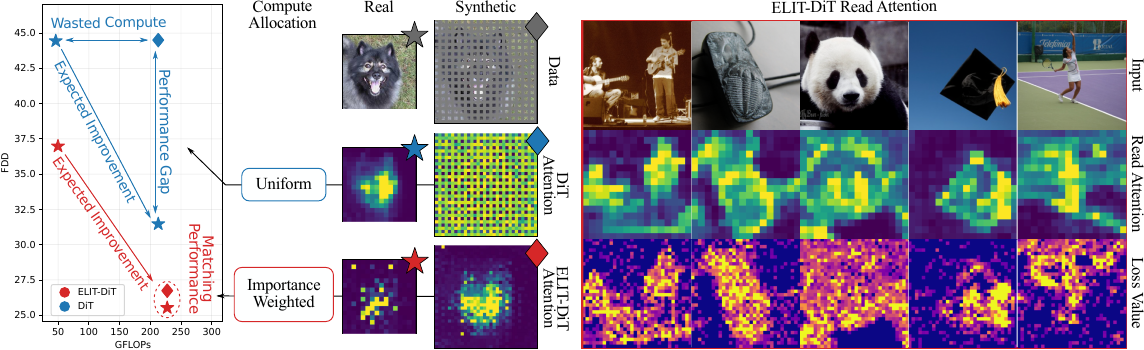}

    \caption{
    \textbf{Adaptive computation.} We test whether DiT and \methodacronym{}-DiT can reallocate compute across image regions by training on synthetic inputs formed by zero-padding real images, artificially increasing the token count (\ding{117}). We compare its performance to baselines trained on real data using patch size $2\!\times\!2$ and patch size $1\!\times\!1$ (\ding{72}).
    Vanilla DiT does not improve: attention in zeroed regions targets other zeroed regions (see “DiT Attention”), so extra tokens raise cost without benefits. In contrast, \methodacronym{}-DiT uses the Read layer to pull informative spatial tokens into the latent interface (see “Read Attention”), effectively filtering out zeroed areas (see “\methodacronym{}-DiT Attention”). Consequently, it leverages the larger token budget and matches the real-data baseline at equal FLOPs. 
  }
  \label{fig:synthetic_experiment_combined}
  \vspace{-1em}
\end{figure*}

Important previous work has sought forms of flexibility to make generative models more efficient. %
Adaptive generators allow a \emph{single} model to operate at different budgets, %
but still spread computation uniformly across tokens~\citep{anagnostidis2025flexidit,yu2018slimmable} or suffer from high complexity~\citep{zhao2025dyditplusplus}.
Masking-based methods improve training speed by skipping tokens. %
Yet, dropping is disabled for inference to avoid unrecoverable information loss~\citep{gao2023maskeddiffusiontransformer,krause2025tread}, making inference compute unchanged.  %
On the other hand, some training-free acceleration approaches reduce the computation for the least informative tokens during inference, leaving the training efficacy the same~\citep{liu2025ras,yang2025sparse,jeong2025ralu}. 
A complementary thread moves flexibility in the autoencoder by learning variable-length representations but stops short of endowing a \emph{generative} model with an internal flexible representation~\citep{bachmann2025flextok,li2025adaptok}. %
Finally, RINs~\citep{rin, chen2023fit} learn to distribute computation non-uniformly across input tokens through a set of latent tokens, but keep inference budget fixed and depart significantly from the DiT architecture, hindering adoption. 

Building on the previous observations, we propose \method{} (\methodacronym{}) (see \Cref{fig:teaser}), a minimal, DiT-compatible mechanism for representation-and-compute adaptivity. We introduce two lightweight cross-attention layers, \emph{Read} and \emph{Write}, that equip DiT-like architectures with a set of latent tokens we refer to as \emph{latent interface}. These latent tokens act as a variable-size surface onto which to distribute input information in a flexible and learnable manner based on the difficulty of each region.
\emph{Read} pulls information from input tokens, which we will refer to as \emph{spatial}, into the latent interface, prioritizing challenging regions. %
\emph{Write} broadcasts updated latents back to the spatial tokens. Importantly, the number of latent tokens is user-controlled, directly setting the per-step compute budget. 
No changes to the training objective are necessary. %

We provide a thorough analysis of \methodacronym{}, showing that it successfully redistributes compute non-uniformly across input tokens across varying base architectures. Our latent interface consistently improves over a fixed-grid model with ImageNet-1k 512px FDD (Fréchet Distance on DINOv2~\citep{dino_v2} features), improving 58.0\% for DiT, 34.0\% for U-ViT, 37.4\% for HDiT. \methodacronym{} also allows for graceful compute-quality tradeoffs by selecting the number of latent tokens at inference time, regularly achieving better tradeoffs than sampling steps reduction while being compatible with training-free acceleration techniques~\citep{liu2024timestep}. %
Additionally, variable compute enables~ autoguidance~\citep{karras2024guiding} out of the box, which reduces inference cost by $\approx 33\%$ without affecting the generation quality. 
In summary, this simple addition
yields a framework capable of: (I) \textbf{Adaptive computation.} Compute is concentrated where it matters rather than spread uniformly across input regions. (II) \textbf{Variable test-time compute.} A single set of weights serves a spectrum of latency-quality points by selecting the number of latent tokens. (III) \textbf{Improved sampling.} Variable compute enables autoguidance~\citep{karras2024guiding} out of the box. (IV) \textbf{Drop-in training.} We keep the design simple, retaining the vanilla rectified-flow objective and showing our method's applicability to DiT, U-ViT, HDiT, and MMDiT. Implementation amounts to only adding the Read and Write layers and latent tokens sampling during training. 

\section{Related Work}
\label{sec:related_work}

\textbf{Adaptive generators for inference budget control.}
Supernetwork trains a single set of weights to support many sub-networks, %
allowing test-time accuracy–efficiency trade-offs~\citep{yu2018slimmable,cai2020onceforall}. Other works train transformers with multiple patchification sizes for variable compute budgets~\citep{anagnostidis2025flexidit,liu2025luminavideo}. %
\citep{zhao2025dyditplusplus} uses learnable routers to adjust network width and drop tokens in MLPs.
These methods differ in \emph{where} adaptivity lives but share the goal of a \emph{single} model that gracefully scales compute at inference time.
Our method adopts a simple variable-length latent interface, improving model convergence and enabling control over the inference budget. %

\noindent\textbf{Token dropping for training speedup.}
Another direction accelerates model training by skipping tokens. MaskDiT~\citep{zheng2024fast} restructures DiTs as an encoder-decoder model and randomly drops encoder input while using an auxiliary reconstruction objective. %
MDTv2~\citep{gao2023maskeddiffusiontransformer} similarly leverages masked latent modeling to train on partial inputs. TREAD~\citep{krause2025tread} randomly selects a set of tokens that skip computation of all blocks from a predefined start to an end DiT block index. 
However, due to the destructive nature of token dropping, such methods typically rely on auxiliary losses~\citep{zheng2024fast}, full-token post-training~\citep{zheng2024fast,krause2025tread}, and adopt full-token inference~\citep{zheng2024fast,gao2023maskeddiffusiontransformer,krause2025tread}, limiting acceleration during inference.
Our method speeds up convergence while enabling control over inference budget by selecting variable amounts of tokens, without applying auxiliary losses.

\noindent\textbf{Latent interfaces.} Latent tokens have been used as compact representations in several architectures. Neural Turing Machines~\citep{graves2014neuralturingmachines} employed them as memory, while Perceiver~\citep{jaegle2021perceiver} used cross-attention to condense high-dimensional inputs. In generative models, RINs~\citep{rin} and FITs~\citep{chen2023fit} introduced interleaved read/write operations for efficient high-dimensional synthesis, which was later scaled to video generation~\citep{snapvideo}. Despite their efficiency, such designs diverge from DiTs and often require specialized optimizers (e.g., LAMB~\citep{lamb}). Similarly, TiTok~\citep{yu2024titok} applied latent tokens as bottlenecks in autoencoders, and recent work extends this to variable-length token sets via tail dropping~\citep{koikeakino2020stochasticbottleneckratelessautoencoder,bachmann2025flextok,li2025adaptok,yan2025elastictok}. Our work brings variable-length latent interfaces to generative models, integrating seamlessly into DiTs with only lightweight Read and Write layers.

\section{Method}
\label{sec:method}
\begin{figure*}
    \centering
    \includegraphics[width=\linewidth]{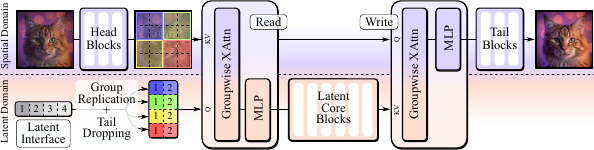}

    \caption{\textbf{Architecture of \methodacronym{}.} We extend a DiT-like generator with a variable-length set of latent tokens—the \emph{latent interface}—and lightweight Read/Write cross-attention layers. A short spatial DiT head processes patchified inputs; Read pulls information into the latent domain where core blocks operate. Write broadcasts updated latents back to spatial tokens, and a small spatial tail produce output. 
    Spatial tokens and latents are partitioned into corresponding groups, with cross-attention operate only within groups.
    During training, we randomly drop tail latents, yielding an importance-ordered interface. At inference, the number of latents serves as a user-controlled compute knob.}

    \label{fig:architecture}
    \vspace{-1em}
\end{figure*}

We propose~\method{} (\methodacronym{}) (see \Cref{fig:architecture}), a novel approach that enables flexible compute allocation in DiT-like transformers. The core component is a variable-length set of latent tokens—the \emph{latent interface}—%
where most transformer blocks operate. Two lightweight cross-attention layers, Read and Write, pass information between domains: Read pulls content from spatial tokens into the latent interface, prioritizing harder regions, while Write broadcasts the updated latent state back to the spatial domain. Unlike the spatial domain, where model FLOPs are a fixed function of input resolution, the latent interface is trained with random tail token dropping, making it resizable. The length of this latent interface during inference directly adjusts FLOPs for each model call. 

\Cref{sec:flow_matching} reviews Flow Matching, \Cref{subsec:rigid_dit} presents a motivating experiment, \Cref{sec:adaptivedit} details the architecture, and \Cref{sec:elastic_computation} describes training and sampling.

\subsection{Preliminaries: Flow Matching}
\label{sec:flow_matching}

We train our generators with the Rectified Flow (RF) variant of flow matching~\citep{liu2022rectifiedflow,lipman2023flowmatching}, which learns a deterministic velocity field connecting a noise distribution $\noisedistribution$ to the data distribution $\datadistribution$. Let $\inputtensorclean \sim \datadistribution$ and $\inputtensornoise \sim \noisedistribution=\mathcal{N}(0,\mathbf{I})$. A linear path is defined as 
$\inputtensor_{\difftimestep} \;=\; (1-\difftimestep)\,\inputtensornoise + \difftimestep\,\inputtensorclean,
 \difftimestep \in [0,1],$
whose ground‑truth velocity is constant along the path:
$\velocity_{\difftimestep} \!=\! \frac{d\inputtensor_{\difftimestep}}{d\difftimestep} \!=\! \inputtensorclean-\inputtensornoise$.
A neural network $\generator(\cdot)$ predicts the velocity from a noised input and time and is optimized as:
{\small
\begin{equation}
\label{eq:rf}
\mathcal{L}_{\text{RF}}
= \mathbb{E}_{\difftimestep \sim \timestepdistribution,\;\inputtensorclean \sim \datadistribution,\;\inputtensornoise \sim \noisedistribution}
\big\|\,\generator(\inputtensor_{\difftimestep},\,\difftimestep) - (\inputtensorclean-\inputtensornoise)\,\big\|_2^2,
\end{equation}
}
where $\timestepdistribution$ is a logit‑normal~\citep{sd3} training distribution over $\difftimestep$. At inference, samples are obtained by integrating the learned ODE from $\inputtensornoise$ to $\inputtensorclean$ with a standard solver. 

\subsection{Uniform Spatial Computation in DiTs}
\label{subsec:rigid_dit}
Standard DiTs operate in the spatial domain where an input $\inputtensortime$ at time $\difftimestep$ is patchified by a linear projection layer into $\numtokens$ tokens and processed by $\numblocks$ transformer blocks. Each block output is connected to the previous block through a \emph{residual connection} at the same spatial location,
 maintaining a fixed mapping between tokens $\spatialact$ in intermediate blocks and corresponding spatial location of $\inputtensortime$. This results in a uniform compute distribution across spatial locations in $\inputtensortime$. 

We probe this behavior through an experiment presented in \Cref{fig:synthetic_experiment_combined}. We train DiT-B/2 (i.e., $2\!\times\!2$ patchification) and a corresponding DiT-B/1 (i.e., $1\!\times\!1$ patchification) on ImageNet-1k. As expected, performance improves in DiT-B/1 due to the fourfold amount of tokens. We then introduce a synthetic ImageNet-1k version by padding encoded images with zeros, yielding four times more tokens.  We train a DiT-B/2, named DiT-B/2-Synth, without loss on padded regions.
We also use absolute learnable positional encodings to avoid bias toward zero patches. 
For evaluation, we remove padded regions before decoding to recover real image content. DiT-B/2-Synth matches the number of tokens and FLOPs of DiT-B/1. Thus, if compute were used effectively, it should match DiT-B/1 performance. Instead, as shown in \Cref{fig:synthetic_experiment_combined}, DiT-B/2-Synth matches DiT-B/2 in FID, indicating no benefit from the added compute.

Attention maps in \Cref{fig:synthetic_experiment_combined} reveal that in DiT-B/2-Synth, zeroed tokens primarily attend to each other instead of important image regions, wasting compute.
We conclude that DiT cannot reallocate computation from easy to hard regions. 
Such inefficiency likely also arises in natural images, where spatial regions vary in difficulty (lower or higher losses) and would benefit from compute redistribution. %

\subsection{Elastic Latent Interface Transformer}
\label{sec:adaptivedit}

\noindent\textbf{From spatial tokens to a variable latent interface.} To allow flexible distribution of computation in DiTs, we introduce a minimal change that eliminates the fixed mapping between tokens and image patches, as shown in \Cref{fig:architecture}. We create a latent domain by instantiating a \emph{latent interface} of $\latenttokcount$ tokens.
To map the original spatial domain to the new latent domain, we use a lightweight \emph{Read} cross‑attention layer, following~\cite{chen2023fit}, which enables the model to select the number of latent tokens adaptively for each spatial region of $\inputtensortime$, based on their difficulty. This forms a compact latent domain on which most transformer blocks operate. 
Finally, a lightweight \emph{Write} cross-attention layer maps the latent updates back to the spatial grid, allowing the model to write predictions back to their locations and retain input details.

\noindent\textbf{Architecture.}
Earlier work has shown that early and late transformer blocks in diffusion models exhibit different specializations compared to the intermediate blocks~\citep{chen2024accelerating, delta_dit, ddt}. Therefore, we split  transformer blocks into three segments:
\newcommand{\improv}[1]{\textsuperscript{\textcolor{green!60!black}{#1}}}

\begin{table*}[t]
\centering
\small
\setlength{\tabcolsep}{2pt}

\newcolumntype{,}{!{\color{gray!35}\vrule}}
\resizebox{\linewidth}{!}{
\begin{tabular}{
  @{}l C{1.3cm} %
  @{\hspace{4pt}} , @{\hspace{4pt}}
  C{1.0cm} c c c c c c  %
  @{\hspace{8pt}}
  C{1.0cm} c c c c c c  %
  @{}
}
\toprule

\multirow{3}{*}{{Model}} & \multirow{3}{*}{{Params}} &
\multicolumn{7}{c|}{{ImageNet 256${\times}$256}} &
\multicolumn{7}{c}{{ImageNet 512${\times}$512}} \\
\cmidrule(lr){3-9}\cmidrule(lr){10-16}
& &
{TF} & \multicolumn{2}{c}{{$\fidfifk\downarrow$}} & \multicolumn{2}{c}{{$\fddfifk\downarrow$}} & \multicolumn{2}{c|}{{IS$\uparrow$}} &
{TF} & \multicolumn{2}{c}{{$\fidfifk\downarrow$}} & \multicolumn{2}{c}{{$\fddfifk\downarrow$}} & \multicolumn{2}{c}{{IS$\uparrow$}} \\
& &
{\scriptsize @256} & {\scriptsize –G} & {\scriptsize +G} & {\scriptsize –G} & {\scriptsize +G} & {\scriptsize –G} & {\scriptsize +G} &
{\scriptsize @512} & {\scriptsize –G} & {\scriptsize +G} & {\scriptsize –G} & {\scriptsize +G} & {\scriptsize –G} & {\scriptsize +G} \\
\midrule

\rowcolor{gray!10}
\textbf{DiT-XL} & 675M & 182 & \cellthird{13.0} & \cellthird{5.7} & \cellthird{346.3} & \cellthird{232.9} & \cellthird{66.2} & \cellthird{115.3} & 806 & \cellthird{18.8} & \cellthird{9.5} & \cellthird{339.2} & \cellthird{233.6} & \cellthird{53.0} & \cellthird{86.4} \\
\ourmodel \hspace{1em}$\llcorner$ \methodacronym{} & 698M & 188 & \cellsecond{8.2} & \cellfirst{3.8} & \cellfirst{200.2} & \cellfirst{124.5} & \cellsecond{93.0} & \cellsecond{160.1} & 831 & \cellsecond{11.1} & \cellsecond{4.9} & \cellsecond{175.6} & \cellsecond{106.1} & \cellsecond{80.0} & \cellsecond{134.1} \\
\ourmodel \hspace{2em}$\llcorner$ \textit{MB} & 698M & 190
& \cellfirst{7.8}\improv{-40\%} & \cellfirst{3.8}\improv{-33\%}
& \cellsecond{203.7}\improv{-41\%} & \cellsecond{128.6}\improv{-45\%}
& \cellfirst{99.0}\improv{+50\%} & \cellfirst{167.6}\improv{+45\%}
& 804
& \cellfirst{10.1}\improv{-46\%} & \cellfirst{4.5}\improv{-53\%}
& \cellfirst{164.1}\improv{-52\%} & \cellfirst{98.2}\improv{-58\%}
& \cellfirst{88.8}\improv{+68\%} & \cellfirst{147.0}\improv{+70\%} \\
\addlinespace[2pt]

\rowcolor{gray!10}
\textbf{UViT-XL} & 707M & 196 & \cellthird{8.3} & \cellsecond{3.8} & \cellthird{220.2} & \cellthird{138.0} & \cellthird{84.4} & \cellthird{145.1} & 861 & \cellthird{11.6} & \cellthird{5.3} & \cellthird{202.7} & \cellthird{125.9} & \cellthird{72.5} & \cellthird{117.2} \\
\ourmodel \hspace{1em}$\llcorner$ \methodacronym{} & 730M & 202 & \cellsecond{7.5} & \cellsecond{3.8} & \cellsecond{203.8} & \cellfirst{130.0} & \cellsecond{95.2} & \cellsecond{159.7} & 886 & \cellsecond{8.9} & \cellsecond{4.2} & \cellsecond{155.3} & \cellsecond{94.9} & \cellsecond{85.8} & \cellsecond{141.0} \\
\ourmodel \hspace{2em}$\llcorner$ \textit{MB} & 730M & 204
& \cellfirst{7.1}\improv{-14\%} & \cellfirst{3.7}\improv{-3\%}
& \cellfirst{203.2}\improv{-8\%} & \cellsecond{130.6}\improv{-5\%}
& \cellfirst{100.3}\improv{+19\%} & \cellfirst{168.2}\improv{+16\%}
& 858
& \cellfirst{7.7}\improv{-34\%} & \cellfirst{3.8}\improv{-28\%}
& \cellfirst{135.8}\improv{-33\%} & \cellfirst{83.1}\improv{-34\%}
& \cellfirst{98.0}\improv{+35\%} & \cellfirst{159.3}\improv{+36\%} \\
\addlinespace[2pt]

\rowcolor{gray!10}
\textbf{HDiT-XL} & 1.4B & 182 & 12.8 & 5.6 & 361.6 & 247.0 & 68.7 & 119.7 & 776 & 13.0 & 6.0 & 260.3 & 170.5 & 69.4 & 114.2 \\
\ourmodel \hspace{1em}$\llcorner$ \methodacronym{} & 1.4B & 188 & \cellfirst{9.4} & \cellfirst{4.6} & \cellsecond{272.2} & \cellfirst{184.4} & \cellsecond{89.5} & \cellsecond{150.2} & 801 & \cellsecond{10.1} & \cellfirst{4.6} & \cellfirst{164.1} & \cellsecond{112.0} & \cellsecond{88.8} & \cellsecond{141.0} \\
\ourmodel \hspace{2em}$\llcorner$ \textit{MB} & 1.4B & 191
& \cellfirst{9.4}\improv{-27\%} & \cellfirst{4.6}\improv{-18\%}
& \cellfirst{271.8}\improv{-25\%} & \cellsecond{185.0}\improv{-25\%}
& \cellfirst{92.3}\improv{+34\%} & \cellfirst{155.7}\improv{+30\%}
& 791
& \cellfirst{9.6}\improv{-26\%} & \cellfirst{4.6}\improv{-23\%}
& \cellsecond{171.2}\improv{-34\%} & \cellfirst{106.8}\improv{-37\%}
& \cellfirst{94.7}\improv{+36\%} & \cellfirst{154.6}\improv{+35\%} \\
\bottomrule
\end{tabular}
}
\caption{
\textbf{Comparative performance on ImageNet-1K at 256px and 512px resolutions.}
We evaluate FID$\downarrow$, FDD$\downarrow$, and IS$\uparrow$ without (–G) and with 0.25 CFG (+G). 
TFLOPs (TF) indicate single training iteration TFLOPs. Superscripts show percentage of improvement 
of \methodacronym{} MultiBudget (MB) relative to the baseline.
}
\label{tab:final_results_v4}
\end{table*}

\vspace{-1mm}
\begin{enumerate}[leftmargin=2.4em,labelsep=0.5em,itemsep=0.0em,topsep=0.0em]

\item\textbf{Short spatial head} ($\numblocksstart$ blocks). Processes input tokens $\spatialact \!\in\! \mathbb{R}^{\numtokens \times d}$ to produce a refined spatial representation which is transferred to the latent interface. This avoids reading from raw noisy patches.

\item\textbf{Latent core} ($\numblocksmid$ blocks). Variable‑length latent sequence $\latentact \in \mathbb{R}^{\latenttokcount \times d}$ drives most computation. We insert a Read layer $\readlayer$ that pulls information from spatial tokens into $\latentact$, then process $\latentact$ with standard transformer blocks in the latent domain, and finally insert a Write layer $\writelayer$ that broadcasts updated latent information back to the spatial domain.

\item\textbf{Short spatial tail} ($\numblocksend$ blocks). A few blocks complete processing the written features and project them to the output velocity. This head restores fine spatial detail, noise information, and aligns features to the prediction space of $\velocitypred$.
\end{enumerate}

\noindent\textbf{Read and Write layers.}
Let $\spatialact \in \mathbb{R}^{\numtokens \times d}$ be the current spatial tokens after the spatial head and $\latentact \in \mathbb{R}^{\latenttokcount \times d}$ a learnable set of initial latent tokens. The Read layer updates the latent interface via cross‑attention from spatial tokens, producing output latents $\latentactout \in \mathbb{R}^{\latenttokcount \times d}$ as follows:
{\small
\begin{equation}
\label{eq:read}
\begin{aligned}
\latentactca &=
  \latentact + \mathrm{CA}\!\bigl(\mathrm{Queries}=\latentact;\,
  \mathrm{Keys,Values}=\spatialact\bigr),\\
\latentactout &=
  \latentactca + \mathrm{MLP}(\latentactca).
\end{aligned}
\end{equation}
}
Conversely, the Write layer updates the spatial representation with the results of latent computations, producing updated spatial tokens $\spatialactout \in \mathbb{R}^{\numtokens \times d}$ and is fully symmetric. %
We adopt pre‑norm, and use adaLN-Zero~\citep{dit} modulation for Read to keep the interface timestep‑aware. To improve stability, we apply $QK$ normalization inside cross‑attention operations. No hidden dimensionality expansion is applied to the MLP blocks to reduce the computational overhead.

\noindent\textbf{Grouped cross‑attention.}
To reduce the cost of Read and Write operations, we partition spatial tokens into $\numgroups$ non‑overlapping groups (e.g., a regular 2D/3D grid for images/videos), and latents are partitioned accordingly in groups of $\latenttokcountgroup=\latenttokcount/\numgroups$ latent tokens each. %
Latents are initialized from a set of learnable positional embeddings, which is reused across groups and encodes positional information \emph{within} each group. 
This removes any dependency on a fixed input resolution: increasing spatial resolution modifies $\numgroups$ and $\numtokens$, but not the number of learnable latents $\latenttokcountgroup$.
Cross attention operations attend \emph{within} corresponding groups only. This turns the cross‑attention cost from $\mathcal{O}(\numtokens\latenttokcount)$ into $\mathcal{O}(\numtokens\latenttokcount/\numgroups)$~\citep{chen2023fit}.

\subsection{Elastic Computation with \methodacronym{}}
\label{sec:elastic_computation}

\noindent\textbf{Spatial compute redistribution.} When integrated with DiT (\methodacronym{}-DiT), our architecture enables spatial compute redistribution. %
In the synthetic dataset experiment described in~\Cref{subsec:rigid_dit}, \methodacronym{}-DiT-B/2-Synth repurposes the compute from zeroed regions to enhance generation in real regions, matching the quality of the baseline trained only on the original ImageNet-1k with equivalent compute(\methodacronym{}-DiT-B/1) (see \Cref{fig:synthetic_experiment_combined}). Attention maps of the read operations averaged over all latent tokens confirm this behavior, showing that \methodacronym{}-DiT builds its latent representation by focusing on the most informative spatial regions with the highest flow-matching loss. %
\begin{figure}[t]

    \begin{minipage}{\linewidth}
        \centering
        \includegraphics[width=0.7\linewidth]{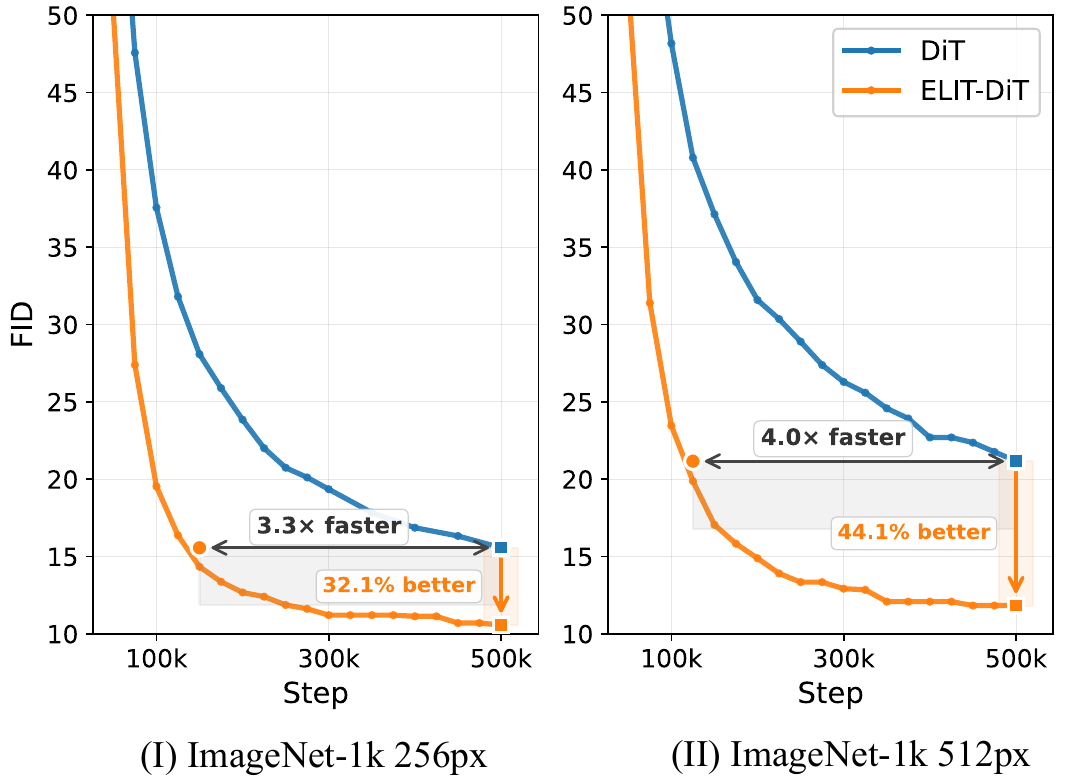}
        \caption{\textbf{Training convergence.} \methodacronym{}-DiT significantly accelerates convergence, achieving \(3.3\times\) speedup on ImageNet-1K 256px and \(4.0\times\) on ImageNet-1K 512px.}
        \label{fig:convergance}
        \vspace{-1em}
    \end{minipage}
    \hfill %

    \hfill %
\vspace{-2em}
\end{figure}

\noindent\textbf{Multi-budget elastic latent interface.} 
We aim to build a multi-budget model that supports variable inference budgets. Since each latent token summarizes its group information via the read operation, a subset of tokens can still predict for the entire group, enabling budget-adaptive inference. Thus, we train an importance-ordered latent interface, where earlier tokens within each group capture globally useful information and later tokens refine details, so that any prefix of $\latenttokcountgroupmasked\!<=\!\latenttokcountgroup$ tokens of the group tokens yields a valid interface associated with reduced computation (see \supp~\Cref{appx:compute_analysis}). We enforce this with a simple random prefix‑keeping scheme as in \cite{rippel2014learningordered}. At training time, we randomly sample $\latenttokcountgroupmasked\!\sim\!\mathrm{Uniform}\{\latenttokcountgroupmin,\ldots,\latenttokcountgroupmax\}$ once per training iteration, defining the training budget for the current iteration. The same value of $\latenttokcountgroupmasked$ is used across all groups. In every Read/Write and latent‑core block, we keep only the first $\latenttokcountgroupmasked$ latents of each group and drop the subsequent tail. %
This process creates a consistent hierarchy where head latents are seen (and trained on) more often, forcing the model into storing the most important information in earlier latents. The generator is trained end‑to‑end only with the standard RF loss in \Cref{eq:rf}. %

\noindent\textbf{Asymmetric compute for improved guidance.} Classifier-free guidance (CFG)~\citep{ho2022classifierfree} is a cornerstone technique in diffusion model sampling. Given a conditioning signal $\conditioning$ and guidance scale $\lambda$, CFG redefines the velocity prediction %
$\velocitypred_{\text{CFG}}
= (\lambda\!+\!1) \generator(\inputtensortime \mid \conditioning)
\!-\!\lambda\generator(\inputtensortime \mid \emptyset)\text{.}$
While improving the quality, this comes at the cost of duplicating the number of model invocations. Recently, AutoGuidance (AG)~\citep{karras2024guiding} was proposed to improve guidance by replacing the unconditional model with a weak version of itself. 
While producing consistent improvements, it relies on the availability of a weaker model version, ideally producing artifacts that are similar to the main model~\citep{chen2025s2guidancestochasticselfguidance}. Weak models are separately trained or obtained through handcrafted model corruptions~\citep{chen2025s2guidancestochasticselfguidance,hong2024smoothed,hyung2025spatiotemporalskipguidance}. %
However, in our multi-budget framework, such model is natively available by varying the inference budget defined by $\latenttokcountgroupmasked$.
Thus, we evaluate the main term with budget $\latenttokcountgroupmasked$ and the guidance term with a smaller budget $\latenttokcountgroupweak\!\le\!\latenttokcountgroupmasked$. We find, however, that AG degrades metrics that reward class alignment, such as Inception Score. We thus opt to drop the class condition from the guidance term, combining the power of AG and CFG, and name the resulting guidance mechanism cheap classifier-free guidance (CCFG). The guidance mechanisms are defined as:

{\small
\begin{equation}
\label{eq:autoguidanceccfg}
\begin{aligned}
\velocitypred_{\text{AG}}
  &= (\lambda+1)\,\generator\Bigl(\inputtensortime \mid \conditioning;\, \latenttokcountgroupmasked\Bigr)
     - \lambda\,\generator\Bigl(\inputtensortime \mid \conditioning;\, \latenttokcountgroupweak\Bigr),\\
\velocitypred_{\text{CCFG}}
  &= (\lambda+1)\,\generator\Bigl(\inputtensortime \mid \conditioning;\, \latenttokcountgroupmasked\Bigr)
     - \lambda\,\generator\Bigl(\inputtensortime \mid \emptyset;\, \latenttokcountgroupweak\Bigr).
\end{aligned}
\end{equation}
}

This results in both improved quality and a reduced cost without any retraining or handcrafted model corruptions.

\section{Experiments}

\subsection{Experimental setup}
\label{sec:experimental_setup}

We demonstrate \methodacronym{}'s broad applicability by evaluating it on several popular diffusion backbones: DiT~\citep{dit}, U-ViT~\citep{bao2023uvit}, and HDiT~\citep{crowson2024hdit}. To ensure a fair comparison and evaluate architectural advantages in isolation, all baselines are built with the \emph{same base transformer blocks}, adopt the same RF framework, and have similar train compute. Furthermore, we integrate several improvements across all baselines, including RoPE~\citep{rope}, and QK normalization~\citep{sd3}.

\noindent\textbf{Training details.}
We perform conditional image and video generation on ImageNet-1k~\citep{deng2009imagenet} and Kinetics-700~\citep{carreira2022shortnotekinetics700human}, respectively. We train on 256px and 512px resolutions for ImageNet-1k experiments and use 29 frames, at 24 fps and 256px resolution for Kinetics-700. We use the FLUX~\citep{flux} autoencoder for images and the CogVideo~\citep{cogvideox} autoencoder for videos. Main experiments %
are based on DiT-XL/2, while ablation studies use a DiT-B/2 model. Unless noted, we use a batch size of 256, learning rate $1\times10^{-4}$ with 10k warmup steps, gradient clipping at 1.0, Adam~\citep{kingma2014adam}, and EMA with $\beta=0.9999$. Image experiments are trained for 500k steps, while video experiments are trained for 200k steps.

\noindent\textbf{Evaluation metrics and protocol.}
We follow the evaluation protocol of~\cite{hajiali2025dfm}. For images, we report FID~\citep{heusel2017gans}, FDD (Fréchet Distance on DINOv2~\citep{dino_v2} features), and Inception Score (IS)~\citep{salimans2016improved}. For video, we report FID, FDD, and FVD~\citep{fvd}. %
Statistics are computed over 50k samples for main image experiments, while 10k samples are used for all other experiments. %
We use an Euler sampler with 40 steps unless otherwise noted. We use FLOPs to measure the amount of computation in all experiments and show in~\supp~
the relationship between FLOPs and forward time.

\subsection{Comparison to baselines}
\label{sec:baselines_comparison}

\begin{figure}
        \begin{minipage}{\linewidth}
        \includegraphics[width=\linewidth]{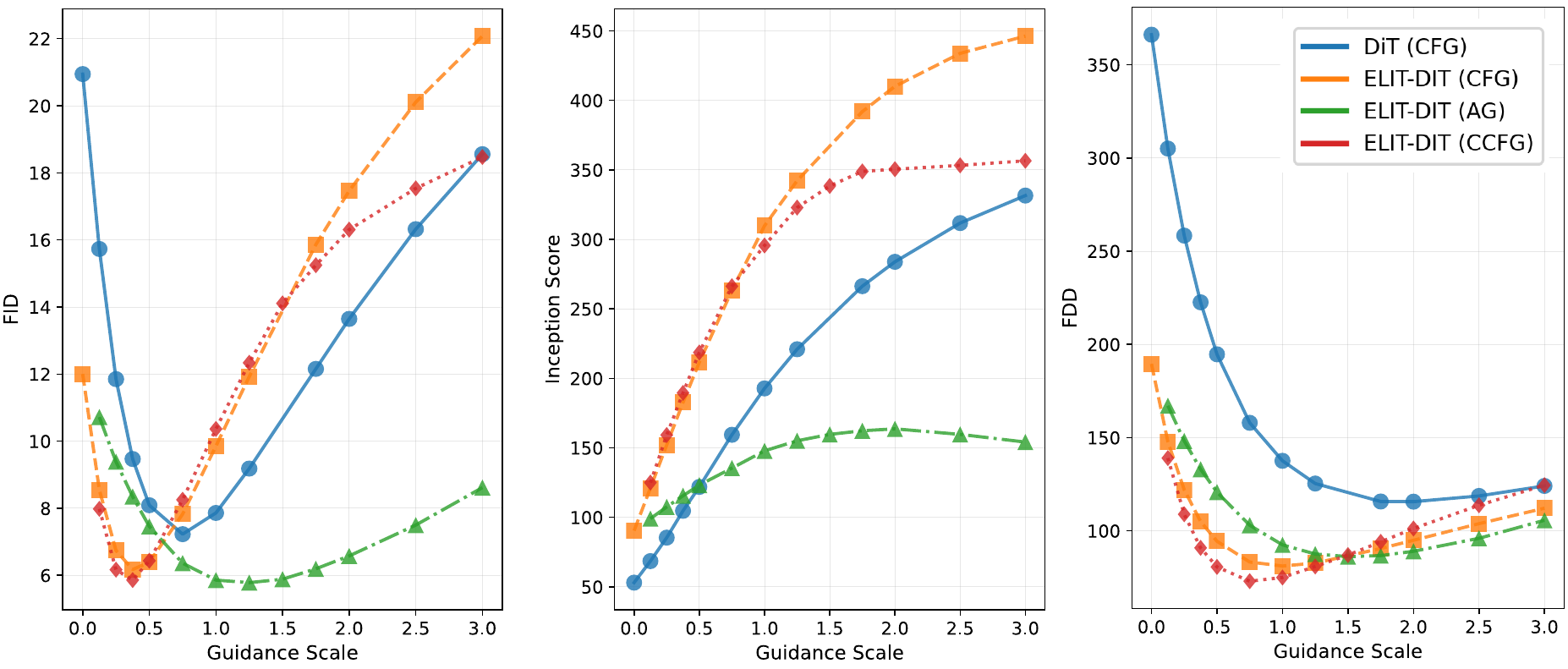}
        \caption{\textbf{Guidance strategies.} \methodacronym{} enables autoguidance  out of the box by providing a well-aligned weaker model that runs at \(\approx 35\%\) of the cost for the unconditional path. When paired with classifier-free guidance (CFG), denoted as cheap CFG (CCFG), it reduces overall generation cost by \(\approx 33\%\) while improving quality. Compared to DiT, \methodacronym{}-DiT achieves a \(19\%\) better best FID.}

        \label{fig:cfg_plot}
    \end{minipage}
    \vspace{-1em}
    
\end{figure}

\begin{figure}
    \centering
    \includegraphics[width=\linewidth]{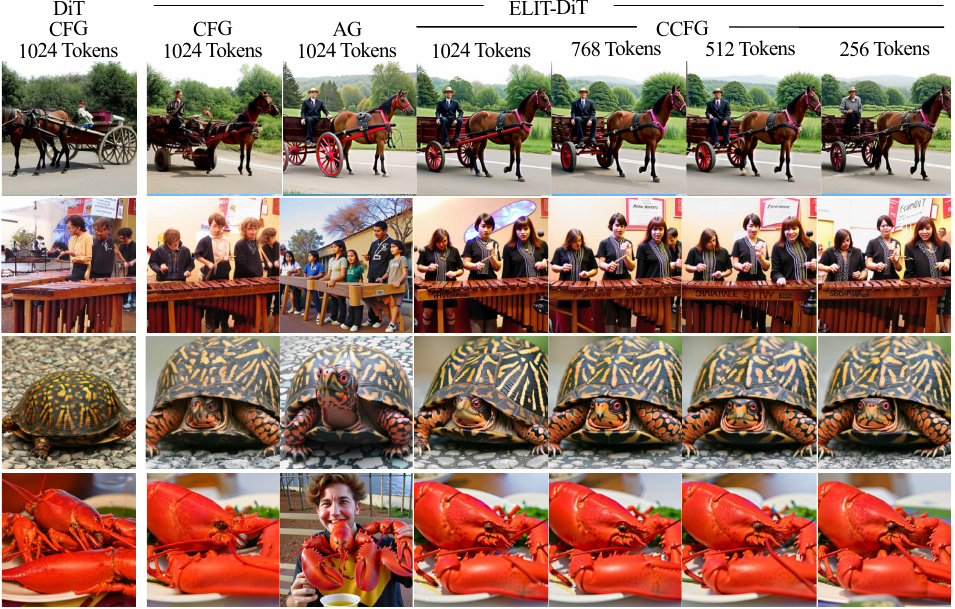}
    \caption{\textbf{Qualitative assessment of \methodacronym{} on ImageNet-1K 512px.} We compare DiT against \methodacronym{}-DiT and ablate over different guidance settings and number of latent tokens. \methodacronym{} improves structural details while allowing per-step selection of inference budget through token dropping and giving access to autoguidance (AG) and cheap classifier free guidance (CFG) for improved sampling quality and cost. See results in \supp~\Cref{appx:additional_evaluation}.} 
    \label{fig:main_qualitatives}
    \vspace{-2em}
\end{figure}

\noindent\textbf{Baseline selection and details.} 
We use DiT~\citep{dit} as a base architecture for our experiments and consider two variants: U-ViT and HDiT. U-ViT~\citep{bao2023uvit} adds long skip connections akin to U-Net~\citep{unet}. We add \methodacronym{} read/write operations while keeping the U-ViT design untouched to obtain \methodacronym{}-U-ViT. HDiT~\citep{crowson2024hdit} reduces tokens in intermediate blocks via PixelUnshuffle/PixelShuffle. We use a single $2{\times}2$ down-/up-sampling operation at blocks 8 and 20, respectively and double the bottleneck hidden dimensionality to match vanilla DiT FLOPs at the cost of an increased parameter count. We apply \methodacronym{} read/write operations and apply group-wise downsampling in the latent space to obtain \methodacronym{}-HDiT. More details are in~\supp~\Cref{supp:baseline_details}.

\noindent\textbf{Image generation.}  To disentangle the effect of dynamic compute redistribution from multi-budget training, we evaluate two variants for each baseline: \methodacronym{}, trained with a single budget matching the baselines configurations, and \methodacronym{}-MB, trained in a multi-budget setup following the tail-token dropping strategy from~\Cref{sec:elastic_computation}. This yields 60 budgets at $512\mathrm{px}$ and 16 budgets at $256\mathrm{px}$. To account for the compute saved at shorter token lengths, we increase the batch size to 384, keeping training FLOPs comparable. We report per-iteration FLOPs for all baselines.

As shown in \Cref{tab:final_results_v4}, \methodacronym{}, despite its simplicity, improves over all baselines under similar training FLOPs. In the multi-budget setting, \methodacronym{}-MB delivers further gains, achieving sizable improvements over DiT, U-ViT, and HDiT across all metrics, with FID reductions of $40\%$, $14\%$, and $27\%$, respectively. We attribute these extra gains to the importance ordering, which leads to better token semantics while benefiting from the larger effective batch size. The improvements are even more pronounced at $512\mathrm{px}$, where FID decreases by $53\%$, $28\%$, and $23\%$ for DiT, U-ViT, and HDiT, respectively, suggesting that our method scales favorably with higher resolution, where pixel redundancy is greater and dynamic compute redistribution is more beneficial. We report in \Cref{fig:convergance} the convergence speed of \methodacronym{}-DiT relative to DiT at both resolutions, showing faster convergence. \Cref{fig:cfg_plot} compares performance across CFG values and confirms the advantage of our method under the optimal CFG scale. Finally, \Cref{fig:main_qualitatives}  shows qualitatively the improvements of our method over DiT. More qualitative examples are provided in \supp~\Cref{appx:additional_evaluation}.

\begin{figure}[t]
    \centering
    \includegraphics[width=\linewidth]{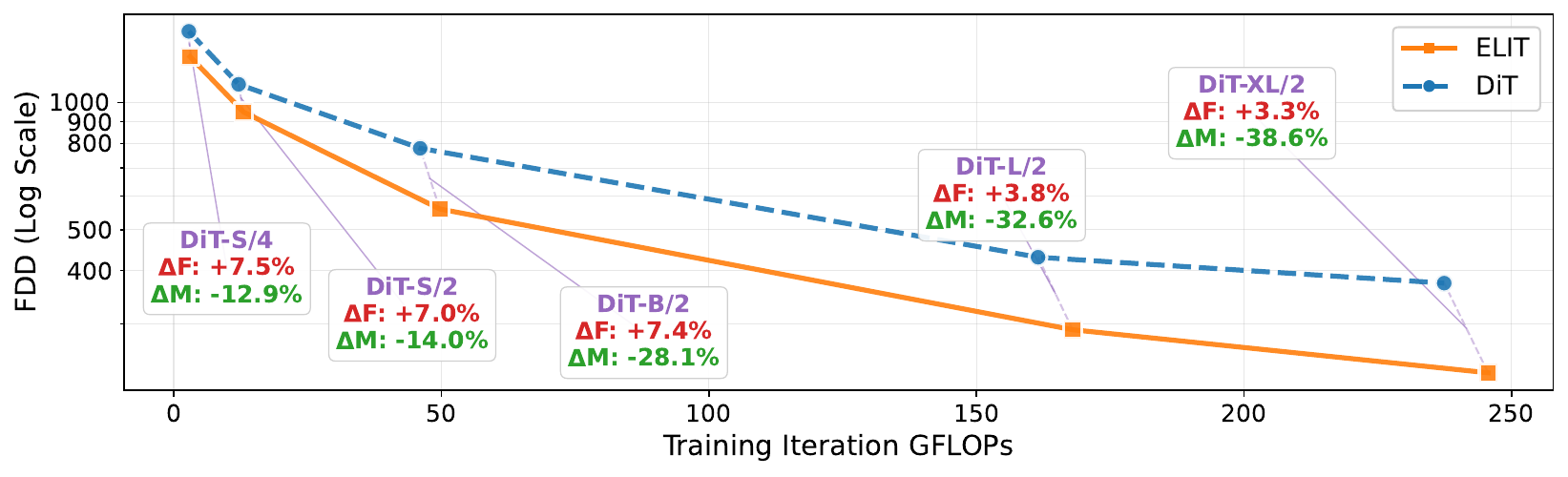}
    \vspace{-2em}
    \caption{\update{\methodacronym{} improves over DiT across model sizes.}}
    \label{fig:model_size}
    \vspace{-1em}
\end{figure}

\noindent\textbf{Video generation.} We validate the performance of \methodacronym{} in class-conditional video generation and report the results in \Cref{tab:kinetics_results} where we apply grouping of spatial and latent tokens in the \emph{frame and temporal dimension}. \methodacronym{}-DiT shows favorable results over DiT across all metrics. 

\update{\noindent\textbf{Scaling across model sizes.} We evaluate \methodacronym{} across model sizes from DiT-S/4 to DiT-XL/2 in \Cref{fig:model_size}. \methodacronym{} outperforms DiT at every scale. Gains are larger for bigger models, while the relative overhead decreases, suggesting that \methodacronym{} is particularly well-suited for large-scale models.}

\begin{figure}[t]
    \centering
    \begin{minipage}{0.45\linewidth}
        \centering
        \includegraphics[width=\linewidth]{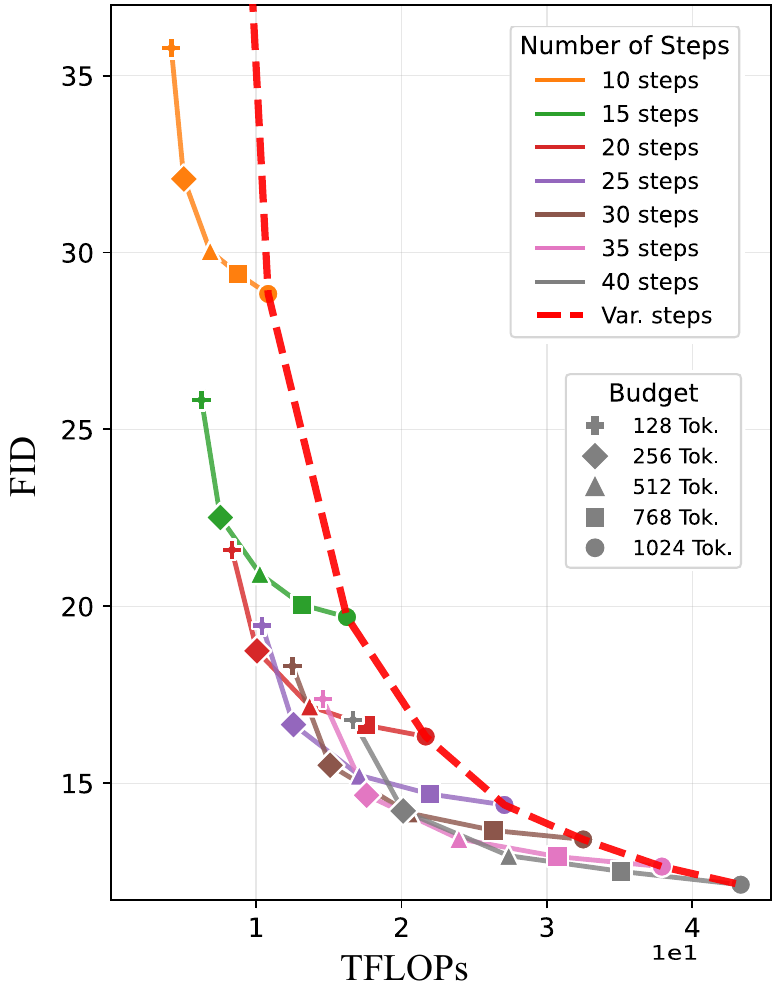}
        \captionof{figure}{ \textbf{Compute-quality tradeoff.} Varying inference budget in \methodacronym{}-DiT give better quality–compute tradeoff than reducing sampling steps.}
        \label{fig:budget_vs_step}
    \end{minipage}
    \hfill
    \begin{minipage}{0.45\linewidth}
        \centering
        \includegraphics[width=\linewidth]{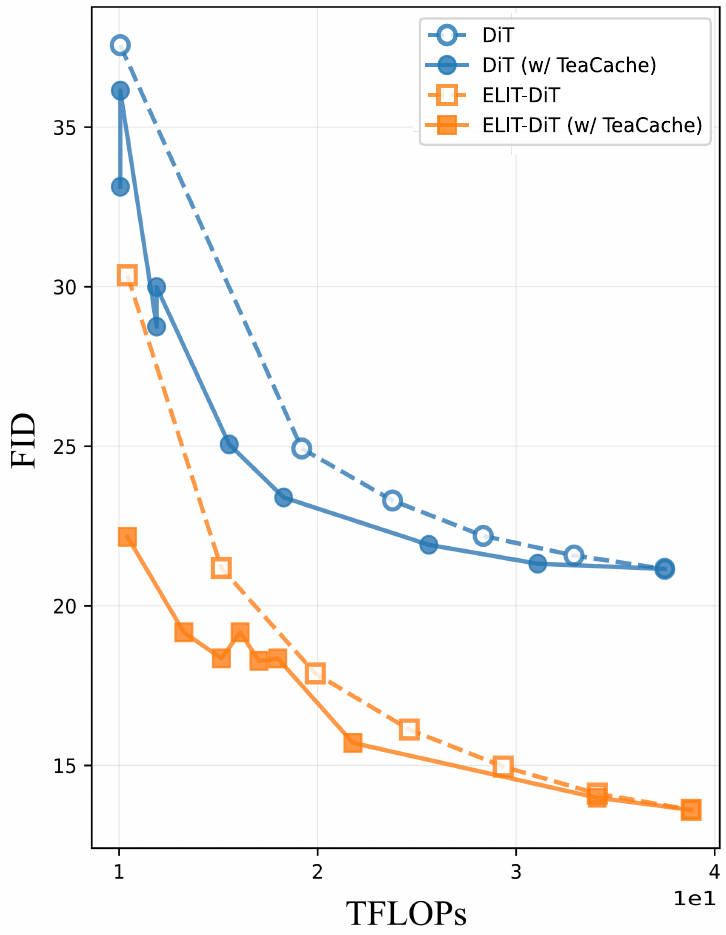}
        \captionof{figure}{ \textbf{TeaCache.} \methodacronym{} benefits from TeaCache similarly to DiT, yielding comparable improvements at different inference FLOPs.}
        \label{fig:teacache}
    \end{minipage}
    \vspace{-1.3em}
\end{figure}

\subsection{Elastic Inference Capabilities}

We analyze the ability of the model to perform inference at varied budgets using the number of retained latent tokens per group $\latenttokcountgroupmasked$ after dropping as a knob to control the budget.

\noindent\textbf{Sampling steps trade-off.} We compare our approach for controlling inference compute against naively lowering the sampling steps.  As shown in \Cref{fig:budget_vs_step}, our method delivers a superior compute–quality trade-off to varying the step count. Notably, for each FLOP target, the optimal combination of number of steps and tokens count varies, underscoring the value of models that support a continuum of inference budgets. We also demonstrate compatibility with TeaCache~\citep{liu2024timestep} in \Cref{fig:teacache}, where our method attains gains comparable to the baseline when TeaCache is applied.

\noindent\textbf{Efficient model guidance.} In \Cref{fig:cfg_plot}, we compare classifier-free guidance (CFG) with autoguidance (AG) and cheap classifier-free guidance (CCFG). \Cref{fig:main_qualitatives} and \supp~\Cref{appx:additional_evaluation}, show qualitative examples of such guidance strategies. AG achieves comparable performance to CFG while using $\approx 33\%$ fewer FLOPs. Combining AG with CFG by dropping the class condition in the guidance term (\textit{i.e.} CCFG) gives the best results in all metrics and delivers the same $\approx 33\%$ inference speedup.

\begin{table}[t]
    \centering
    \small
    \setlength{\tabcolsep}{4.0pt}

    \resizebox{\linewidth}{!}{%
    \begin{tabular}{
        @{}l
        C{1.0cm} C{1.0cm}
        C{1.0cm} C{1.0cm}
        C{1.0cm} C{1.0cm}
        @{}
    }
    \toprule
    \multirow{2}{*}{{Baseline}} & \multicolumn{2}{c}{$\fddtenk\downarrow$} & \multicolumn{2}{c}{$\fidtenk\downarrow$} & \multicolumn{2}{c}{$\fvdtenk\downarrow$} \\
    \cmidrule(lr){2-3}\cmidrule(lr){4-5}\cmidrule(lr){6-7}
    & {--G} & {+G} & {--G} & {+G} & {--G} & {+G} \\
    \midrule
    \rowcolor{gray!10}
    \textbf{DiT-XL} & 371.5 & 309.1 & 14.0 & 11.3 & 135.9 & 100.5 \\
    \addlinespace[2pt]
    \ourmodel \hspace{1em}$\llcorner$ \methodacronym{} & \cellfirst{277.4} & \cellfirst{222.0} & \cellfirst{13.3} & \cellfirst{10.7} & \cellfirst{116.5} & \cellfirst{90.5} \\
    \bottomrule
    \end{tabular}%
    }
    \caption{\textbf{Comparative performance on Kinetics-700 256px.} We report metrics without (–G) and with 0.25 CFG (+G).}
    \label{tab:kinetics_results}
\end{table}

\vspace{-0.4em} %

\begin{table}[t]
\centering
\setlength{\tabcolsep}{3.0pt}
\footnotesize

\resizebox{\linewidth}{!}{%
\begin{tabular}{lcccccccc}
\toprule
Model & Tokens & FLOPs & Entity & Relation & Attribute & Other & Global & Avg. \\
\midrule
\rowcolor{gray!10}
Qwen-Image &  & $1\times$ & 90.51 & 92.21 & 91.03 & 91.34 & 91.70 & 91.27 \\
\methodacronym{}-Qwen-Image: &  &  &  &  &  &  &  &  \\
\rowcolor{gray!10}
\ourmodel \hspace{1em}$\llcorner$100\% Tokens  & 4096 & $0.688\times$ & 90.30 & 92.18 & 88.97 & 90.34 & 89.18 & 90.45 \\
\rowcolor{gray!10}
\ourmodel \hspace{1em}$\llcorner$50\% Tokens   & 2048 & $0.494\times$  & 90.15 & 89.94 & 89.05 & 90.09 & 89.06 & 89.81 \\
\ourmodel \hspace{1em}$\llcorner$25\% Tokens   & 1024 & $0.409\times$ & 89.31 & 91.87 & 89.71 & 88.28 & 84.79 & 89.79 \\
\rowcolor{gray!10}
\ourmodel \hspace{1em}$\llcorner$12.5\% Tokens & 512 & $0.369\times$ & 91.20 & 90.35 & 88.77 & 89.94 & 79.84 & 88.02 \\
\bottomrule
\end{tabular}
}
\caption{Qwen-Image evaluation on DPG-Bench. Results are reported with different inference budgets.}
\label{tab:qwen_dpgbench}
\vspace{-2em}
\end{table}

\begin{table*}[t]
\centering
\footnotesize
\setlength{\tabcolsep}{3pt}

\begin{minipage}[t]{0.33\textwidth}
\centering
\vspace{0pt}
\par\addvspace{2pt}
\resizebox{\linewidth}{!}{
\begin{tabular}{ccccc}
\multicolumn{5}{c}{\emph{(a) Group Size Ablation}} \\
\toprule
{Group Size} & {Groups} & {$\fidtenk$}$\downarrow$ & {$\fddtenk$}$\downarrow$ & {IS}$\uparrow$ \\
\midrule
\multicolumn{5}{c}{{ImageNet-1K 256px (16$\mathbf{\times}$16 tokens)}}\\
\rowcolor{gray!10}
1$\times$1   & 256 & 29.94 & 638.8 & 38.39 \\
2$\times$2   & 64  & \cellfirst{25.48} & \cellsecond{546.8} & \cellsecond{45.66} \\
\rowcolor{gray!10}
4$\times$4   & 16  & \cellsecond{26.53} & \cellfirst{531.8} & \cellfirst{45.95} \\
8$\times$8   & 4   & \cellthird{27.73} & \cellthird{552.5} & \cellthird{44.64} \\
\rowcolor{gray!10}
16$\times$16 & 1   & 30.03 & 599.1 & 43.44 \\
\midrule
\addlinespace[12.5pt]
\toprule
{Group Size} & {Groups} & {$\fidtenk$}$\downarrow$ & {$\fddtenk$}$\downarrow$ & {IS}$\uparrow$ \\
\midrule
\multicolumn{5}{c}{{ImageNet-1K 512px (32$\mathbf{\times}$32 tokens)}}\\
\rowcolor{gray!10}
1$\times$1   & 1024 & 41.67 & 701.6 & 27.23 \\
2$\times$2   & 256  & 34.50 & 604.0 & 33.99 \\
\rowcolor{gray!10}
4$\times$4   & 64   & \cellsecond{31.60} & \cellsecond{540.1} & \cellthird{37.85} \\
8$\times$8   & 16   & \cellfirst{30.86} & \cellfirst{524.6} & \cellfirst{39.48} \\
\rowcolor{gray!10}
16$\times$16 & 4    & \cellthird{31.93} & \cellthird{545.7} & \cellsecond{38.24} \\
\bottomrule
\end{tabular}
}
\end{minipage}
\hfill
\begin{minipage}[t]{0.29\textwidth}
\centering
\vspace{0pt}
\par\addvspace{2pt}
\resizebox{\linewidth}{!}{
\begin{tabular}{cccc}
\multicolumn{4}{c}{\emph{(b) Blocks Allocation Ablation}} \\
\toprule
{Block Alloc.} & {$\fidtenk$}$\downarrow$ & {$\fddtenk$}$\downarrow$ & {IS}$\uparrow$ \\
\midrule
\multicolumn{4}{c}{DiT-B/2} \\
\rowcolor{gray!10}
0-12-0  & 33.84 & 706.4 & 34.49 \\
1-10-1  & 28.55 & 557.5 & 41.82 \\
\rowcolor{gray!10}
2-8-2   & 26.53 & \cellsecond{531.8} & \cellthird{45.95} \\
3-6-3   & \cellsecond{25.37} & \cellfirst{531.0} & \cellsecond{46.19} \\
\rowcolor{gray!10}
4-4-4   & \cellfirst{25.34} & 560.1 & \cellfirst{46.40} \\
5-2-5   & 26.95 & 612.1 & 43.15 \\
\addlinespace[4pt]
\midrule
\multicolumn{4}{c}{{DiT-XL/2}} \\
\rowcolor{gray!10}
0-28-0  & 13.53 & 333.2 & 76.07 \\
2-24-2  & 12.33 & 239.9 & 86.87 \\
\rowcolor{gray!10}
4-20-4  & 11.14 & \cellfirst{229.6} & \cellthird{93.20} \\
6-16-6  & \cellthird{10.84} & \cellsecond{234.8} & \cellsecond{93.59} \\
\rowcolor{gray!10}
8-12-8  & \cellfirst{10.44} & \cellthird{237.3} & \cellfirst{95.16} \\
10-8-10 & \cellsecond{10.80} & 250.1 & 90.11 \\
\bottomrule
\end{tabular}
}
\end{minipage}
\hfill
\begin{minipage}[t]{0.37\textwidth}
\centering
\vspace{0pt}
\par\addvspace{2pt}
\resizebox{\linewidth}{!}{
\begin{tabular}{lccc}
\multicolumn{4}{c}{\emph{(c) Variable Budget Strategy}} \\
\toprule
{Model} & {$\fidtenk$}$\downarrow$ & {$\fddtenk$}$\downarrow$ & {IS}$\uparrow$ \\
\midrule
\rowcolor{gray!10}
DiT & 39.0 & 779.3 & 29.2 \\
\addlinespace[2pt]
\hspace{1em}$\llcorner$ DiT + var. patch size &  57.36 & 991.2 & 20.34 \\
\hspace{2em}$\llcorner$ 25\% Tokens & 85.25 & 1181.9 & 13.06 \\
\addlinespace[2pt]
\rowcolor{gray!10}
\methodacronym{}-DiT  + rand. drop & \cellsecond{27.0} & \cellsecond{540.1} & \cellsecond{46.3} \\
\hspace{1em}$\llcorner$ 25\% Tokens & \cellsecond{38.6} & \cellsecond{718.0} & \cellsecond{34.5} \\
\addlinespace[2pt]
\rowcolor{gray!10}
\methodacronym{}-DiT + tail drop & \cellfirst{26.6} & \cellfirst{536.8} & \cellfirst{47.2} \\
\hspace{1em}$\llcorner$ 25\% Tokens & \cellfirst{36.3} & \cellfirst{682.1} & \cellfirst{36.4} \\

\addlinespace[18.5pt]
\multicolumn{4}{c}{\emph{(d) Batching Strategy}} \\
\midrule
{Model} & {$\fidtenk$}$\downarrow$ & {$\fddtenk$}$\downarrow$ & {IS}$\uparrow$ \\
\midrule
(i) variable batch size & \cellfirst{26.15} & \cellsecond{537.07} & \cellfirst{48.45} \\
\addlinespace[2pt]
(ii) constant batch size & \cellsecond{26.65} & \cellfirst{536.83} & \cellsecond{47.18} \\
\bottomrule
\end{tabular}
}
\end{minipage}
\caption{\textbf{Ablations.}  (a) Read/Write group size. (b) Blocks allocation to head-latent core-tail blocks. (c) Strategies for achieving variable budget inference. (d) Batching strategy for multibudget training. }
\label{tab:all_ablations}
\vspace{-1em}

\end{table*}

\subsection{Large Scale Multi-Budget Model}
We evaluate the applicability of \methodacronym{} to large-scale generative models by applying it on top of Qwen-Image~\citep{wu2025qwenimage}, which is based on a 20B MM-DiT backbone. We insert the Read and Write layers respectively after block 8 and 52. Due to the asymmetric nature of MM-DiTs and small number of text tokens ($\approx$ 300 on average) versus 4096 image tokens at 1024px, we apply \methodacronym{} to the large image tokens stream only. Rather than outperforming the original model, a task which would require access to large-scale curated image datasets and post-training procedures matching the original ones~\cite{wu2025qwenimage}, the experiment aims to demonstrate that \methodacronym{} enables stable training and multi-budget inference for large-scale MM-DiT at high resolution. Therefore, we fine-tune from Qwen-Image in a distillation setting. Specifically, we fine-tune for 60k steps at $512\mathrm{px}$ resolution, using a combination of RF loss and a distillation loss scaled to a similar magnitude. 
We then fine-tune for an additional 60k steps at $1024\mathrm{px}$. We train on real images and synthetic ones generated from FLUX.1-Schnell~\citep{flux} and SDXL~\citep{podell2023sdxl}.

We perform inference using the Euler sampler with 40 steps and CFG of $5.0$ for the original method, while we use the faster CCFG with same weight for \methodacronym{}-Qwen-Image. Additional details are reported in \supp~\Cref{appx:additional_evaluation}.
\Cref{fig:teaser} and \supp~\Cref{fig:appx_qwen_qualitatives} show qualitative results, where \methodacronym{}-Qwen-Image cuts sampling FLOPs by up to $63\%$, achieving $\approx 2.7\times$ speedup while gracefully trading off speed for quality. On DPG-Bench~\citep{hu2024dpgbench}, it maintains strong performance across different inference budgets, with the average score ranging from $90.45$ to $88.02$ at the lowest budget. We include original Qwen-Image results using the same sampling parameters for completeness. With respect to the original Qwen-Image, we observe an initial score gap of $0.82$ average score points in our model.

\subsection{Ablations}

\noindent\textbf{Group sizes.} The latent group size controls how flexibly the interface can attend over spatial tokens, with larger groups enabling more opportunities for non-uniform compute. As shown in \Cref{tab:all_ablations}~(a), dividing the image into 16 groups performs best across 256px and 512px resolutions. Groups of $1{\times}1$ force a rigid one-to-one, spatially aligned mapping, while $16{\times}16$ spans the full $256$px image and underperforms. We hypothesize that using $>\!1$ groups provides useful coarse spatial regularization, while still permitting intra-group compute redistribution.

\noindent\textbf{Blocks allocation.}  
We vary $(\numblocksstart,\numblocksmid,\numblocksend)$ for DiT-B/2 and DiT-XL/2 in \Cref{tab:all_ablations}~(b) (i.e., the head, latent core, and tail blocks count). Optimal results occur when $\approx\!67\%$ and $\approx\!71\%$ of blocks are in the latent core, respectively for DiT-B and DiT-XL, with the rest split between head and tail. We use $4\!-\!20\!-\!4$ for main experiments on DiT-XL.

\noindent\textbf{Alternative variable-budget strategies.} We evaluate other approaches for variable-budget inference in \Cref{tab:all_ablations}~(c). Following \citet{anagnostidis2025flexidit,liu2025luminavideo}, we train a DiT with two patchification layers ($2{\times}2$ and $4{\times}4$) sampled uniformly during training, and set the batch size to 48 to match the baseline’s training FLOPs. In our experiments, this multi-patchification setup underperforms the standard DiT. We also replace tail-dropping strategy with random token dropping and observe a consistent performance drop.

\noindent\textbf{Training Strategy.}  At each training step, we sample a compute budget by choosing the number of latent tokens per group, yielding variable FLOPs, which is lower on average than the baseline. To match baseline compute, we compare two strategies: (i) variable batch size scaled with lower budget, (ii) a constant batch size chosen to match baseline compute in expectation. As shown in \Cref{tab:all_ablations}~(d), both behave similarly, so we use the simpler constant batch size option.

\section{Discussion}

\methodacronym{} introduces a minimal framework that improves compute allocation in DiTs via lightweight Read/Write layers, enabling flexible compute budgets. It consistently improves image and video generation quality across architectures (DiT, U-ViT, HDiT) and resolutions, while enabling efficient compute–quality trade-offs. However, large-scale, from-scratch training benefits remain unverified. Moreover, the proposed CCFG tends to saturate images faster than CFG. Future work can explore training and inference budget schedulers that allocate different budgets across sampling steps, following prior evidence that early sampling steps require less compute~\citep{pyramidalflow,anagnostidis2025flexidit}.

\noindent
{\bf Acknowledgments. } V.O. was partially funded by a gift from Snap Research, funding from the Ken Kennedy Institute at Rice University and NSF Award \#2201710.

{
    \small
    \bibliographystyle{ieeenat_fullname}
    \bibliography{main}
}

\appendix
\clearpage

\part*{Appendix}
\addcontentsline{toc}{part}{Appendix}   %

\begingroup
  \etocsetnexttocdepth{subsection}     %
  \etocsettocstyle{\section*{Appendix Contents}}{}
  \localtableofcontents
\endgroup

\section{Appendix}
\section{Baseline Details}
\label{supp:baseline_details}

\noindent\textbf{DiT setup.} We follow the standard DiT block design and incorporate recent improvements including QK normalization and rotary position embeddings (RoPE). Training hyperparameters match those of~\citet{dit}: batch size $=256$, $12$ transformer blocks for DiT-B and $28$ for DiT-XL. We use patch size of $2{\times}2$ for all experiments. We train all baseline using rectified flow matching objective and use logit-normal distribution for sampling the timesteps.

\noindent\textbf{U-ViT setup.} U-ViT mirrors DiT but adds U-Net–style residual (skip) connections. To isolate architectural effects, we use the same transformer blocks and training hyperparameters as DiT, differing only in the inclusion of these residual connections.

\noindent\textbf{HDiT setup.} HDiT follows DiT but applies PixelShuffle/PixelUnshuffle to reduce the token count while increasing channel dimensionality. We adopt this token–channel trade-off on the same transformer blocks as baselines. We use a single downsampling/upsampling operation after blocks 6 and 22. We also exclude local attention and instead use full self-attention. We train with the same hyperparameters as the other baselines.

\noindent\textbf{Qwen-Image setup.} We add \methodacronym{} \emph{Read/Write} layers at blocks 8 and 52 of the 60-layer Qwen-Image backbone. Training uses a weighted sum of RF and distillation losses. The distillation term is scaled by $20\times$ to match the magnitude of the RF loss. We train for 60k steps at $512$px with a global batch size of $1536$, followed by 60k steps at $1024$px with a global batch of $384$. We sample timesteps from a logit-normal distribution and use time shifting of $2.22$ during training and $2.0$ during inference, following ~\citep{wu2025qwenimage}. We do not apply any timestep-aware loss re-weighting. The 
training dataset is a combination of internal real images with synthetic samples generated by FLUX.1-Schnell and Stable Diffusion-XL (with 50/50 ratio). We found that the model converges quickly, but we observe a style bias toward the synthetic data (reduced detail and more saturation relative to original Qwen-Image). For sampling, we use the Euler ODE sampler with 40 steps and use CFG value of $6.0$

\noindent\textbf{TeaCache setup.} TeaCache proposes two strategies for deciding when to reuse (cache) the previous step’s prediction:
(1) using \emph{timestep-modulated tensor relative error} between current and previous step to predict the accumulative error of caching the current step.
(2) using \emph{timestep-embedding relative error}, which measures the relative change of the timestep embedding itself across steps.

The original paper reports that strategy (1) generally works better. In text-to-image models (e.g., FLUX~\cite{flux}), input tensors are modulated by the timestep embedding, providing access to the timestep-modulated input tensor. In our class-conditional image and video setting, those tensors are additionally modulated by the class signal, preventing access to timestep-modulated tensor. Empirically, on DiT for class-conditional ImageNet, we found that using class-timestep modulated input tensor following strategy (1) does not provide good estimate for the caching error and leads to degraded quality, underperforming the second strategy. Consequently, we adopt the timestep-embedding relative error (strategy 2) for all TeaCache experiments in this work.

\begin{figure*}[t]
  \centering
    
  \begin{subfigure}[t]{0.6\linewidth}
    \centering
    \setlength{\tabcolsep}{3.0pt}
    \footnotesize
    \adjustbox{valign=t}{
    \resizebox{\linewidth}{!}{%
    \begin{tabular}{lcccc}
      \toprule
       & {Spat. Blocks} & {Lat. Blocks} & {Read} & {Write} \\
      \midrule
       Attn. Proj. & $8\numtokens\hidsize^2$ & $8\latenttokcountgroup\numgroups\hidsize^2$ & $\hidsize^2(4\numtokens\!+\!4\latenttokcountgroup\numgroups)$ & $\hidsize^2(4\numtokens\!+\!4\latenttokcountgroup\numgroups)$ \\
       Attn. Mat. & $2\numtokens^2\hidsize$ & $2\latenttokcountgroup^2\numgroups^2\hidsize$ & $2\latenttokcountgroup\numtokens\hidsize$ & $2\latenttokcountgroup\numtokens\hidsize$ \\
       FF & $16\numtokens\hidsize^2$ & $16\latenttokcountgroup\numgroups\hidsize^2$ & $4\latenttokcountgroup\numgroups\hidsize^2$ & $4\numtokens\hidsize^2$ \\
      \bottomrule
    \end{tabular}
    }}
    \label{tab:model_compute}
  \end{subfigure}
  \hfill
  \begin{subfigure}[t]{0.39\linewidth}
    \centering
    \adjustbox{valign=t}{
    \includegraphics[width=\linewidth]{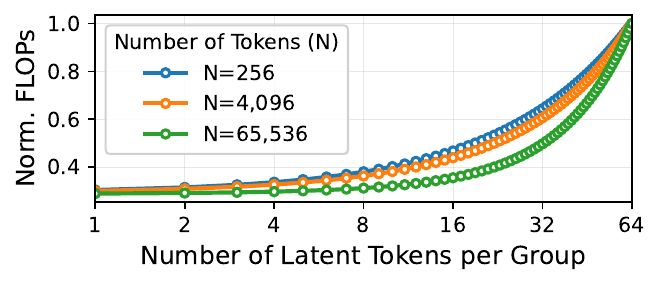}
    }
    \label{fig:compute_vs_latents}
  \end{subfigure}

  \caption{(left) FLOPs for spatial blocks, latent blocks, and Read/Write layers as a function of input tokens $\numtokens$, groups count $\numgroups$, latent tokens per group $\latenttokcountgroup$, and hidden size $\hidsize$. 
  (right) Relationship between latent tokens per group and model FLOPs for a DiT-XL with 8 spatial blocks, 20 latent core blocks, and $\numtokens/64$ groups, varying input tokens $\numtokens$ and latent tokens per group $\latenttokcountgroupmasked$. FLOPs are shown relative to 64 tokens per group. }
  \label{fig:compute_table_combo}
\end{figure*}

\section{Method Details}
\label{supp:method_details}
\noindent\textbf{Adapting \methodacronym{} to baselines.} Aside from adding the Read/Write operations, we leave each baseline’s architecture and training unchanged. Unless noted, we place the \emph{Read} at block 4 and the \emph{Write} at block 24 for XL-size models across all baseliness (DiT, U-ViT, HDiT), as motivated by our ablations in \Cref{tab:all_ablations}.

\noindent\textbf{Multi-budget training setup.} We use 16 spatial groups per image in all main experiments. On ImageNet-1K, each group contains 16 latent tokens at $256$px and 64 at $512$px. Unless otherwise noted, during training we set $\latenttokcountgroupmax$ to the per-group maximum (64 at $512$px; 16 at $256$px) and $\latenttokcountgroupmin$ to $1$ for $256$px and $4$ for $512$px, yielding 16 distinct inference budgets at $256$px and 60 at $512$px. At each training iteration, $\latenttokcountgroupmasked$ is sampled once and broadcast to all GPUs, ensuring synchronized compute with no added overhead. To account for the reduced compute, we increase the batch size from 256 (baselines) to 384 to match training FLOPs.

\noindent\textbf{Kinetics-700 setup.} We train at $256$px on 29 frames sampled at 24 fps. The encoder produces $8$ latent frames of shape $8{\times}32{\times}32$. We use a patch size of $1{\times}2{\times}2$, this yields 2{,}048 tokens. We use a group size of $2{\times}4{\times}4$, giving 64 groups per video. Kinetics-700 is trained with a single compute budget without multi-budget training.

\noindent\textbf{Inference setup.} We use the Euler ODE sampler with 40 steps for all experiments. Image experiments are evaluated on the ImageNet-1K validation split, and video experiments on the Kinetics-700 validation split.

\begin{table*}[t]
    \centering
    \small
    \setlength{\tabcolsep}{3.5pt} %
    \caption{\textbf{Compute-quality tradeoff efficiency of baselines on ImageNet-1K 512px.} $\tradeoffeff=(\mathrm{Metric~Ratio})/(\mathrm{FLOPs~Ratio})$ indicates the model degradation with respect to change in FLOPs between the low- and high-compute variants. } 
    \label{tab:tradeoff_efficiency}
    \resizebox{\linewidth}{!}{
    \begin{tabular}{
        @{}l %
        C{1.1cm} %
        C{1.1cm} %
        C{1.5cm} C{1.5cm} %
        C{1.8cm} C{1.8cm} %
        C{1.8cm} C{1.8cm} %
        @{}
    }
    \toprule
    \multirow{2}{*}{{Baseline}} & \multirow{2}{*}{{Params}} & \multirow{2}{*}{{TFLOPs}} & \multicolumn{2}{c}{{$\fidfifk$}$\downarrow$ ($\tradeoffeff$$\downarrow$)} & \multicolumn{2}{c}{{$\fddfifk$}$\downarrow$ ($\tradeoffeff$$\downarrow$)} & \multicolumn{2}{c}{{IS}$\uparrow$ ($\tradeoffeff$$\downarrow$)} \\
    \cmidrule(lr){4-5}\cmidrule(lr){6-7}\cmidrule(lr){8-9}
    & & & {--G} & {+G} & {--G} & {+G} & {--G} & {+G} \\
    \midrule
    \rowcolor{gray!10}
    \textbf{DiT} & 675M & 806 & 18.8 (1.00) & 9.5 (1.00) &  339.2 (1.00) & 233.6 (1.00) & 53.0 (1.00) & 86.4 (1.00) \\
    \hspace{1em}$\llcorner$ Patch size 2x4 & 675M & 377 & 22.5 (0.56) & 12.3 (0.61) & 434.0 (0.60) & 317.9 (0.74) & 45.7 (0.54) & 73.8 (0.55) \\
    \addlinespace[2pt]

    \textbf{HDiT} &1.4B & 776 & 13.0 (1.00) & 6.0 (1.00) & 260.3 (1.00) & 170.5 (1.00) & 69.4 (1.00) & 114.2 (1.00)\\
    \hspace{1em}$\llcorner$ Smaller backbone & 703M & 392 & 22.2 (0.85) & 11.5 (0.96) & 435.2 (0.83) & 315.4 (0.93) & 48.8 (0.71) & 80.0 (0.71)\\

    \rowcolor{gray!10}
    \textbf{\methodacronym{}-DiT} & 698M & 831 &  11.1 (1.00) & 4.9 (1.00) & 175.6 (1.00) & 106.1 (1.00) & 80.0 (1.00) & 134.1 (1.00) \\
    \ourmodel \hspace{1em}$\llcorner$ 25\% Tok. & 698M & 386 & 12.5 (\cellfirst{0.52}) & 5.7 (\cellfirst{0.54}) & 217.7 (\cellfirst{0.57}) & 137.8 (\cellfirst{0.60}) &75.7 (\cellfirst{0.49}) & 124.5 (\cellfirst{0.50}) \\
    \bottomrule
    \end{tabular}
    }
\end{table*}

\section{Compute-quality Tradeoff Efficiency}

Increasing the training image resolution scales the required compute quadratically, making higher resolution training expensive. To control the compute while keeping model configuration the same, DiT proposed to increase the patch size to cut token count, while HDiT inserts a downsampling stage that reduces tokens but increases parameters count. We instead propose to cap the number of latent tokens per group during training, reducing training compute while keeping both patch size and model size constant.

To evaluate compute–quality trade-offs, we train low/high-compute variants for each baseline: DiT (larger patch size for the low variant), HDiT (model size matching other baselines), and \methodacronym{}-DiT (fewer latent tokens). Intuitively, given a similar reduction in compute between the two versions, the architecture with least performance degradation is more desirable.

\begin{figure} %
    \centering
    \includegraphics[width=0.4\textwidth]{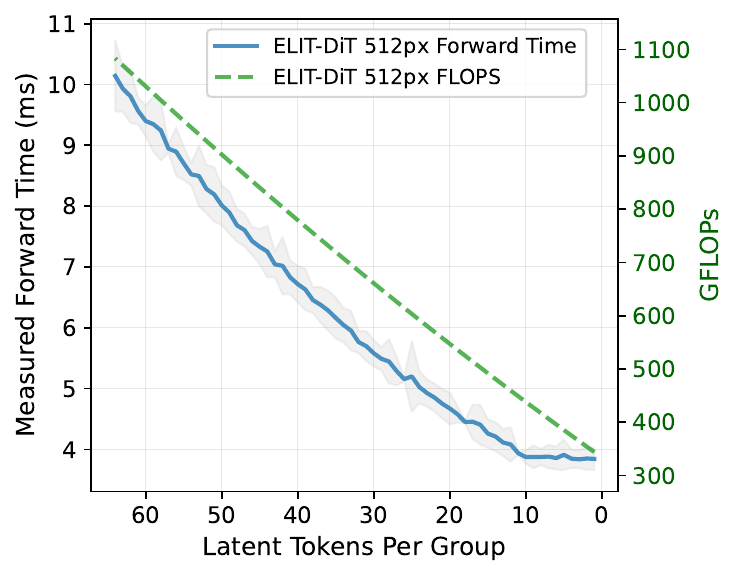}
    \caption{Lowering inference budget by using fewer latent tokens per group yields correlated reductions in forward time and FLOPs.}
    \label{fig:latency}
\end{figure}
 To measure this, we define a degradation metric $\tradeoffeff=((\mathrm{Metric~Ratio})/\mathrm{FLOPs~Ratio})$, where ``Metric Ratio'' represents metric degradation caused by the low-compute model and ``FLOPs Ratio'' represents the corresponding reduction in FLOPs. As shown in \Cref{tab:tradeoff_efficiency}, not only our method outperforms baselines at similar training compute, but also shows consistently lower $\tradeoffeff$ indicating it can more efficiently make use of its comupte if constrained, a capability we attribute to the latent interface's focus on the most important information in the input.

\begin{table}[t]
    \centering

    \resizebox{0.8\linewidth}{!}{%
    \begin{tabular}{cccc}
        \toprule
        {Baseline} & {$\fidtenk$}$\downarrow$ & {$\fddtenk$}$\downarrow$ & {IS}$\uparrow$ \\
        \midrule
        \rowcolor{gray!10}
        \methodacronym{}-DIT & 26.53 & 531.8 & 45.95 \\
        Qformer Read & 30.49 & 589.9 & 41.10 \\
        \rowcolor{gray!10}
        Self-Attn Read  & 28.38 & 602.5 & 40.12 \\
        Self-Attn Read/Write  & 29.46 & 631.1 & 38.49 \\
        
        \rowcolor{gray!10}
        $\uparrow$ Read Capacity   & 27.45 & 540.7 & 45.40 \\
        $\uparrow$ Write Capacity   & \cellsecond{25.23} & \cellsecond{516.9} & \cellsecond{47.59} \\
        \rowcolor{gray!10}
        $\uparrow$ FFN Capacity   & \cellfirst{24.80} & \cellfirst{507.7} & \cellfirst{48.22} \\
        \bottomrule
    \end{tabular}%
    }
            \captionof{table}{\textbf{Architectural ablations on DiT-B/2.} Using cross-attn in Read/Write is superior to alternatives. Increasing the model capacity is only beneficial in Write and FFN.}
            \label{tab:architecture_ablation}
            \vspace{-1em}
\end{table}

\section{Ablations on Read/Write Strategies.} In \Cref{tab:architecture_ablation}, we compare alternative Read/Write designs and find that a single cross-attention Read layer outperforms both a Q-Former–style Read layer~\citep{li2023blip2} and full self-attention. Additionally, stacking two cross-attention layers in the Read yields no measurable gain, suggesting one layer suffices. However, adding a second cross-attention layer in the Write or expanding the FFN hidden dimension by $\times4$ (as in the DiT block) offers improvements at the cost of additional FLOPs. To keep overhead at a minimum, we adopt a single Read/Write layer.

\section{Compatibility with Distillation Methods.} \update{We evaluate the compatibility of \methodacronym{} with the distillation technique such as grafting~\cite{grafting}, which distills a base model into a smaller version of itself. We apply grafting on $100\%$ \methodacronym{} MLPs with expansion ratio $r\!=\!3$, obtaining $12.6\%$ degradation in FID and $8.9\%$ in IS, consistent with the original paper's reported degradation of $17.2\%$ FID and $9.4\%$ IS. This confirms that \methodacronym{} remains compatible with orthogonal efficiency methods such as network pruning and distillation.}

\begin{table}[t]
    \centering
    \caption{\update{Budget scheduling across noise levels. ImageNet 512px, \methodacronym{}-DiT-XL/2. $50\%\_100\%$: uses $50\%$ of tokens for high-noise steps ($t\!<\!0.5$), $100\%$ otherwise.}}
    \label{tab:supp_training_strategy}
    \begin{tabular}{lccc}
    \toprule
    \textbf{Method} & $\textbf{FID}_{10\text{K}} \downarrow$ & \textbf{IS$\uparrow$} & $\text{Iter}_{\text{FLOPs}}$\\
    \midrule
    $100\%\_100\%$ & \textbf{11.60} & 86.68 & 188 \\
    $50\%\_100\%$ & 11.98 & \textbf{90.18} & 154 \\
    \bottomrule
    \end{tabular}
\end{table}

\section{Budget scheduling Across Noise Levels.} \update{We explore allocating different token budgets across noise levels. As a proof of concept, we train \methodacronym{} on ImageNet 512px (DiT-XL/2) with $50\%$ of tokens for high-noise steps ($t\!<\!0.5$) and $100\%$ for the remaining steps ($50\%\_100\%$). As shown in \Cref{tab:supp_training_strategy}, despite lower per-iteration compute (154 vs.\ 188 TFLOPs), performance remains comparable, suggesting that high-noise steps may require fewer tokens. We leave a principled study of noise-level-aware budget scheduling for training and inference as future work.}

\begin{table}[t]
    \centering
    \caption{\update{Joint multi-budget vs.\ independently trained single-budget models. When tested on ImageNet 512px, \methodacronym{}-DiT-XL/2, joint multi-budget models consistently outperform single budget models.}}
    \label{tab:supp_indep_models}
    \begin{tabular}{lccc}
    \toprule
    \textbf{Method} & $\textbf{FID}_{10\text{K}}\downarrow$ & $\textbf{FDD}_{10\text{K}}\downarrow$ & \textbf{IS$\uparrow$} \\
    \midrule
    Indep. (100\% tok.) & 13.60 & 205.23 & 80.90 \\
    Joint  (100\% tok.) & \textbf{12.00} & \textbf{189.50} & \textbf{90.29} \\
    \midrule
    Indep. (50\% tok.)  & 14.14 & 222.43 & 77.99 \\
    Joint  (50\% tok.)  & \textbf{12.95} & \textbf{203.58} & \textbf{85.18} \\
    \midrule
    Indep. (25\% tok.)  & 15.36 & 247.77 & 74.04 \\
    Joint  (25\% tok.)  & \textbf{14.21} & \textbf{228.08} & \textbf{79.60} \\
    \bottomrule
    \end{tabular}
\end{table}

\section{Joint vs.\ Independent Budget Training.} \update{We compare our joint multi-budget model against independently trained single-budget \methodacronym{} models on ImageNet 512px (DiT-XL/2). As shown in \Cref{tab:supp_indep_models}, the joint model consistently outperforms independently trained models across all token budgets ($100\%$, $50\%$, $25\%$) and all metrics. This demonstrates that multi-budget training acts as a regularizer, and that a single \methodacronym{} model natively supporting multiple budgets eliminates the need to train separate models for each budget.}

\begin{figure}[t]
    \centering
    \includegraphics[width=\linewidth]{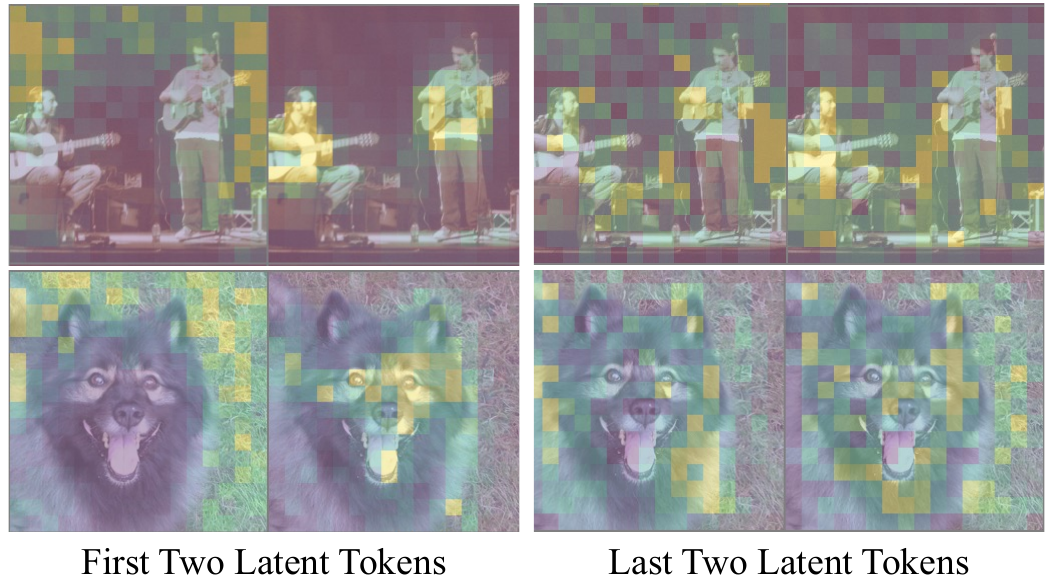}
    \caption{\update{Read attention masks averaged over noise levels. Early latent tokens attend to broad, semantically important image regions, while later tokens exhibit sparser attention focusing on fine-grained details.}}
    \label{fig:supp_latent_tokens}
\end{figure}

\section{Latent Token Importance Visualization.} \update{We visualize the attention mask of the Read operation, averaged over noise levels, in \Cref{fig:supp_latent_tokens}. Early latent tokens attend to broad image regions covering both background structure and the main object, whereas later tokens exhibit sparser attention patterns, often concentrating on fine-grained texture details. This confirms the importance ordering learned through tail dropping and is consistent with the observation that increasing the token count for Qwen-Image primarily improves high-frequency texture details while preserving overall structure (\Cref{fig:teaser}).}

\begin{figure}[t]
    \centering
    \includegraphics[width=\linewidth]{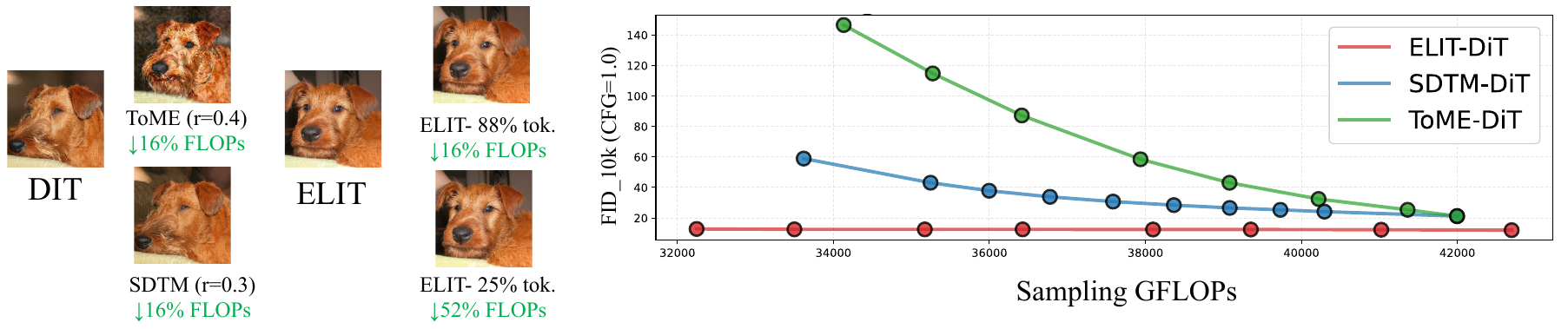}
    \vspace{-0.5em}
    \caption{\update{\methodacronym{} vs.\ token merging methods on ImageNet 512px (DiT-XL/2). Training-free methods (ToMe, SDTM) are bounded by DiT quality, while \methodacronym{} surpasses it even at $25\%$ tokens.}}
    \label{fig:merging_methods}
\end{figure}

\section{Comparison with Token Merging Methods.} \update{Token-merging methods~\cite{sparsedit, toma, tome, stdm, adaptor, Importance-based} can provide a knob to control inference budget. They are often training-free or require lightweight finetuning~\cite{sparsedit, vandchali2025one} We compare \methodacronym{} against training-free token merging approaches (ToMe~\cite{tome}, SDTM~\cite{stdm}) on ImageNet 512px (DiT-XL/2). As shown in \Cref{fig:merging_methods}, both training-free methods trade compute for quality less favorably than \methodacronym{}. \methodacronym{} improves over the base DiT even when using only $25\%$ of the tokens ($\fidtenk\!=\!14.2$), while training-free methods are upper bounded by DiT quality ($\fidtenk\!=\!20.9$).}

\section{Compute Analysis of \methodacronym{}}
\label{appx:compute_analysis}

We analyze the theoretical computation requirement for \methodacronym{}-DiT in comparison with standard DiT design.  
Figure~\ref{fig:compute_table_combo} (left) shows the relation between main architecture hyperparameters and FLOPs for the blocks employed by our architecture. When the number of core blocks is large with respect to spatial blocks, computation is focused on the latent core blocks and the Read and Write operations' cost is minimal with respect to the model cost. %
Figure~\ref{fig:compute_table_combo} (right) exemplifies the case of a DiT-XL/2 architecture %
for varying input sequence lengths. The latent interface is particularly effective at reducing FLOPs with large sequence lengths (\textit{e.g.} training on higher resolutions) due to the dominant self attention cost that is quadratically reduced with $\latenttokcountgroupmasked$.

\noindent\textbf{FLOPs vs latency in \methodacronym{}.} \Cref{fig:latency} reports FLOPs and wall-clock forward 
 time for \methodacronym{}–DiT on ImageNet-1k at $512$px as we vary the number of latent tokens per group. Forward time drops monotonically with token count and closely follow the FLOPs reduction, showing that budget control yields real speedups. At higher budgets, the correlation weakens slightly due to fixed overheads (e.g., I/O and kernel launch), but the overall trend remains strongly aligned.

\section{Additional Results}
\label{appx:additional_evaluation}

\begin{figure}
    \centering
    \includegraphics[width=0.8\linewidth]{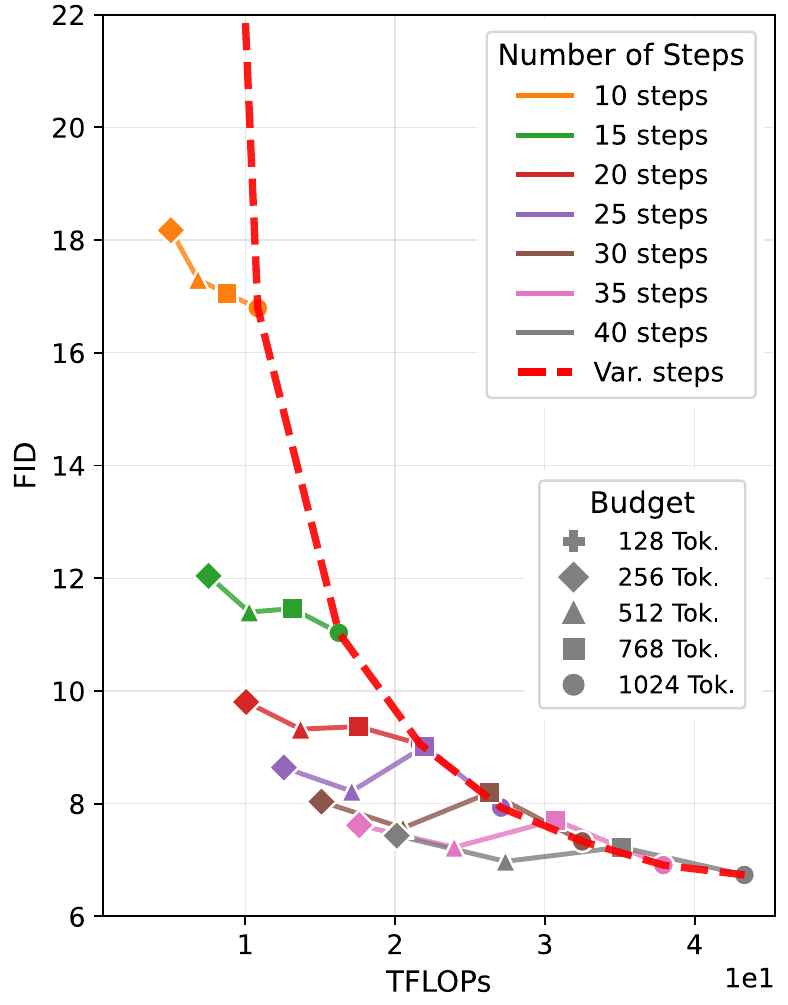}
    \caption{When tested with CFG 0.25, \methodacronym{} provides better quality-compute tradeoff than reducing the number of sampling steps.}
    \label{fig:budget_vs_step_w_cfg}
\end{figure}

\noindent\textbf{Compute-quality tradeoff.} To verify the advantage of our method over simply reducing the number of sampling steps, we show in \Cref{fig:budget_vs_step_w_cfg} that our multi-budget model achieves a more favorable quality–compute tradeoff compared to varying the number of sampling steps.

\begin{figure*}[t]
    \centering
    \includegraphics[width=0.8\linewidth]{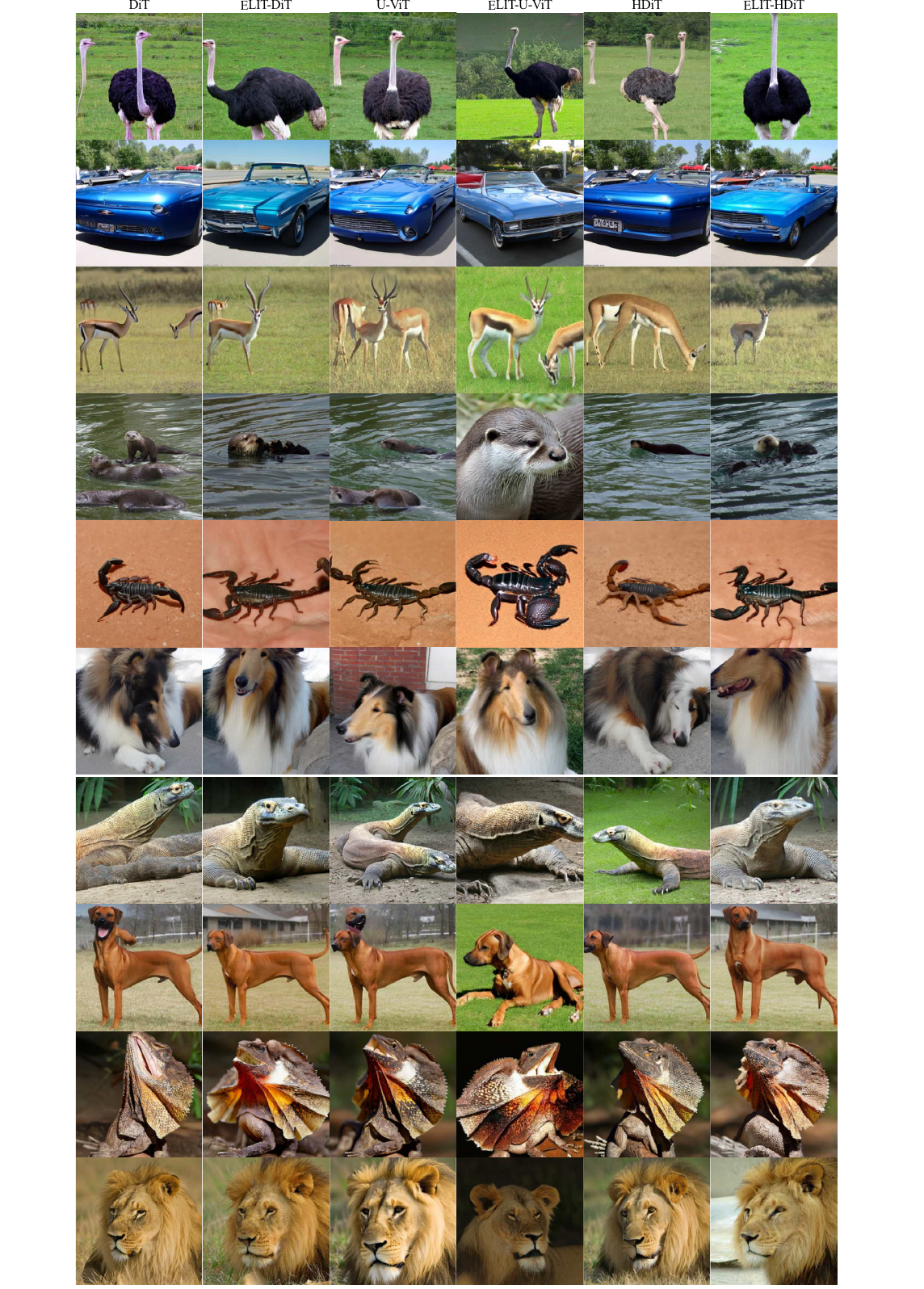}
    \caption{Qualitative comparison of \methodacronym{} against baselines on the ImageNet-1K 512px dataset. Results are produced using CFG with weight 4.0 for all methods.} 
    \label{fig:appx_baseline_comparison_512px}
\end{figure*}

\noindent\textbf{Comparison to baselines.} We show in \Cref{fig:appx_baseline_comparison_512px} additional qualitative results comparing our method to baselines on ImageNet-1K 512px. \methodacronym{} variants show less structural artifacts while allowing for per-step selection of inference budget and enabling autoguidance and cheap classifier-free guidance out of the box for cheaper and higher quality sampling.

\begin{figure*}[t]
    \centering
    \includegraphics[width=0.8\linewidth]{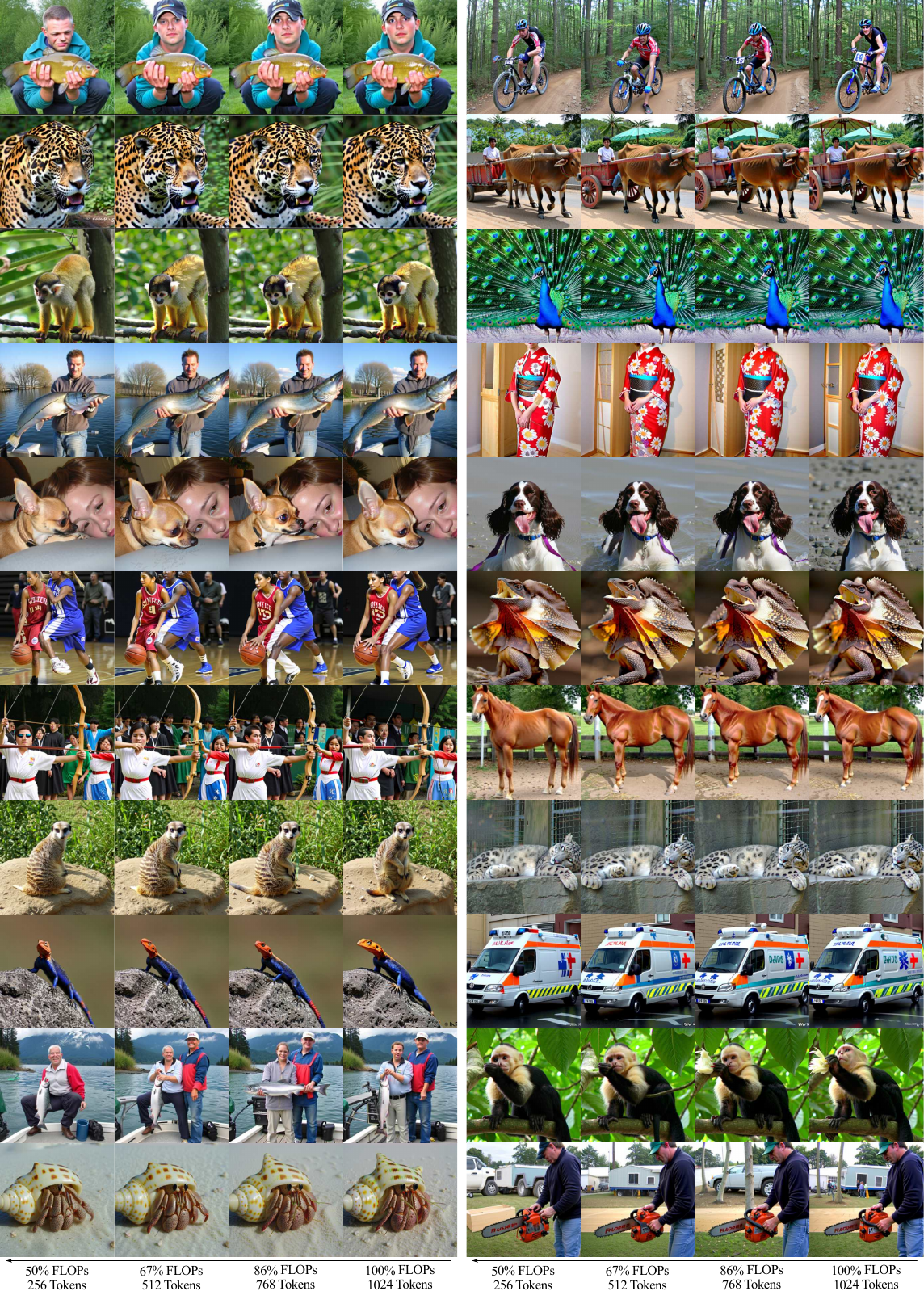}
    \caption{Qualitative results produced by \methodacronym{}-DiT on ImageNet-1K 512px with CCFG 4.0 for varying number of tokens in the latent interface. As the tokens and model FLOPs are reduced, the model preserves structure, while varying image details, producing gradual image changes. FLOPs are expressed relative to the model variant where no latent tokens are dropped.} 
    \label{fig:appx_variable_budgets}
\end{figure*}

\noindent\textbf{Varying inference budget.} In \Cref{fig:appx_variable_budgets}, we evaluate the effects of varying the number of tokens in the latent interface for \methodacronym{}-DiT trained on ImageNet-1K 512px. As the model FLOPs decrease with the number of latent tokens, the model is able to preserve image structure while changing less noticeable details.

\begin{figure*}[t]
    \centering
    \includegraphics[width=0.8\linewidth]{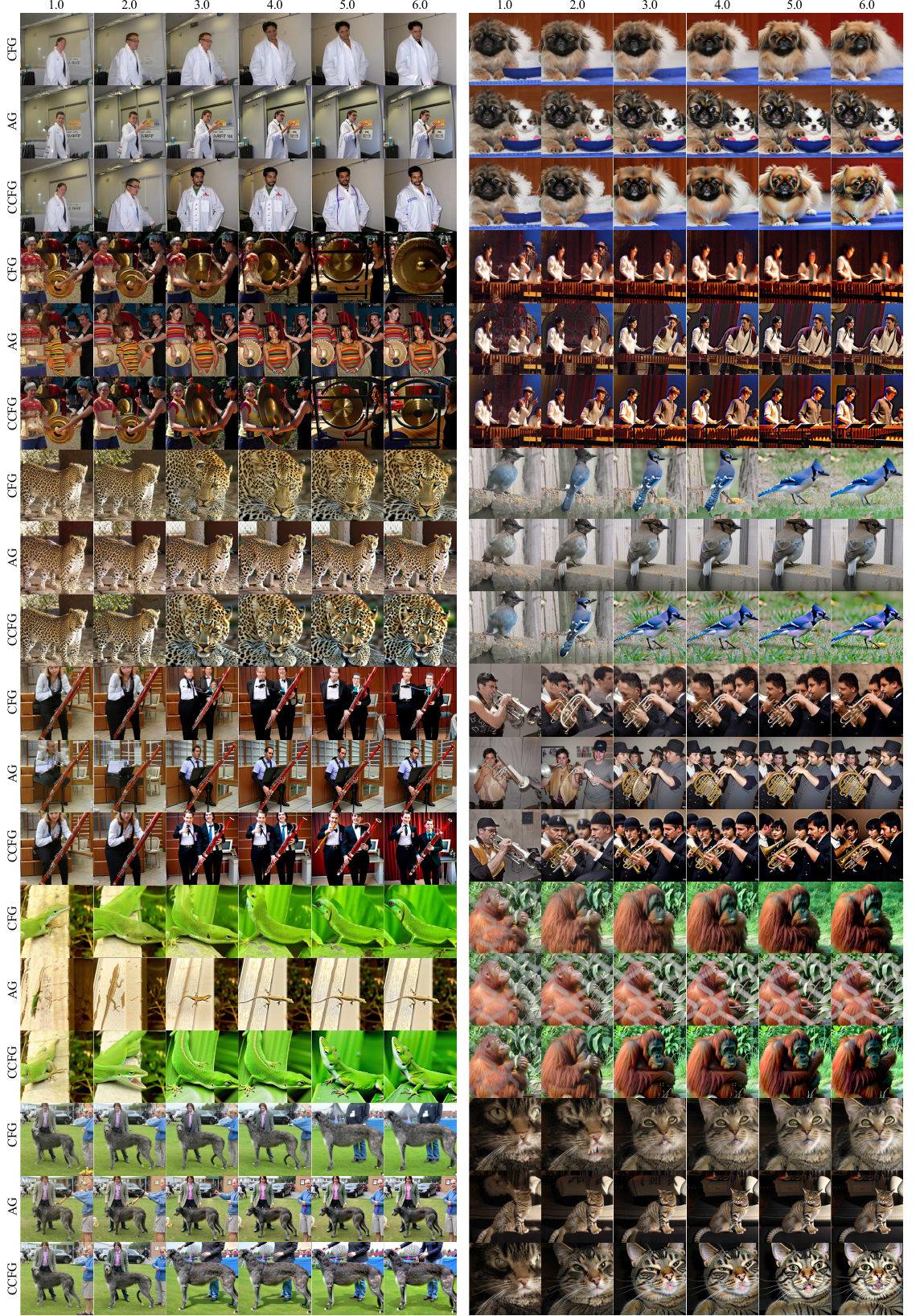}
    \caption{Qualitative comparison of classifier-free guidance (CFG), autoguidance~\citep{karras2024guiding} (AG), cheap classifier-free guidance (CCFG) with different weights, when applied to \methodacronym{}-DiT trained on the ImageNet-1K 512px dataset. AG produces the most varied samples, generating results with similar structure across guidance weights, as opposed to CFG and CCFG which favor object-centric generations. Both AG and CCFG produce better generations of complex concepts such as human faces.} 
    \label{fig:appx_cfg_comparison_512px}
\end{figure*}

\noindent\textbf{Comparison of guidance methods.} We qualitatively evaluate the effects of classifier-free guidance (CFG), autoguidance~\citep{karras2024guiding}, and the proposed cheap classifier-free guidance (CCFG) (see \Cref{fig:appx_cfg_comparison_512px}). We notice that AG produces results with most variation, including wider ranges of camera poses, compositions with multiple subjects and objects occlusion. By comparing results across different weights, we notice that AG remains most closely aligned with low guidance weight results, avoiding the mode collapse effect visible for CFG and CCFG that pushes samples towards more object-centric representations for the given class. We attribute this observation to the lower Inception Scores obtained by AG in \Cref{fig:cfg_plot}. Both AG and CCFG produce improved results which are particularly noticeable in complex concepts such as humans. CCFG combines the object-centric behavior of CFG, while reaping improved generation of complex objects from AG.

\begin{figure}[t]
    \centering
    \includegraphics[width=0.45\linewidth]{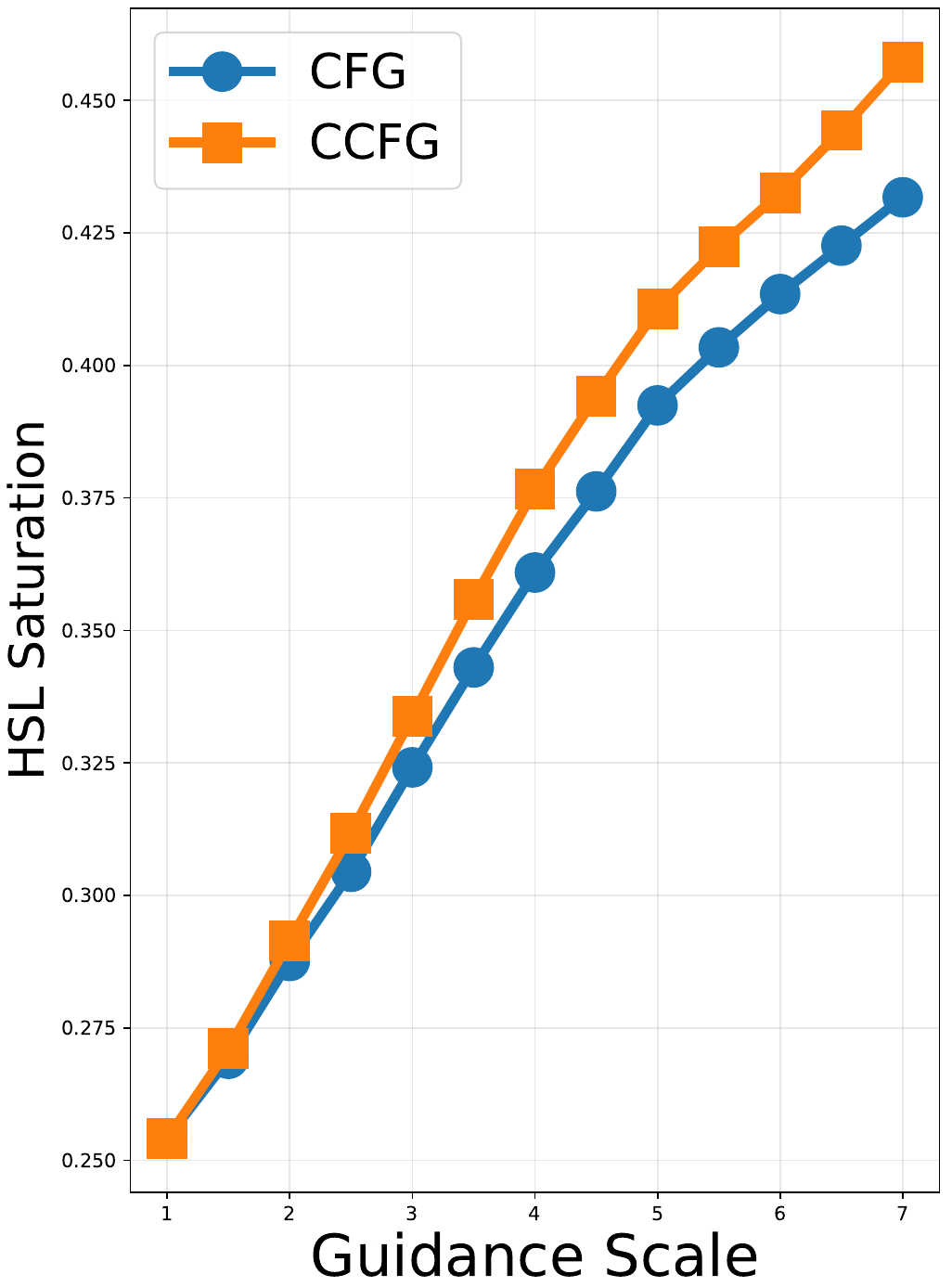}
    \caption{\update{HSL saturation comparison between CFG and CCFG, across guidance scales on ImageNet 512px (DiT-XL/2). CCFG exhibits slightly higher saturation than CFG, attributed to its autoguidance component.}}
    \label{fig:supp_saturation}
\end{figure}

\update{\noindent\textbf{CCFG saturation analysis.} We quantitatively analyze the saturation behavior of CCFG compared to CFG and AG. As shown in \Cref{fig:supp_saturation}, CCFG exhibits slightly higher HSL saturation across guidance scales, which we attribute to the stronger guiding effect contributed by its autoguidance component. The qualitative comparisons in \Cref{fig:appx_cfg_comparison_512px} shows that CCFG tends to saturate at larger guidance scales. Thus, we recommend using lower guidance scales with CCFG to mitigate this effect.}

\begin{figure*}[t]
    \centering
    \includegraphics[width=0.8\linewidth]{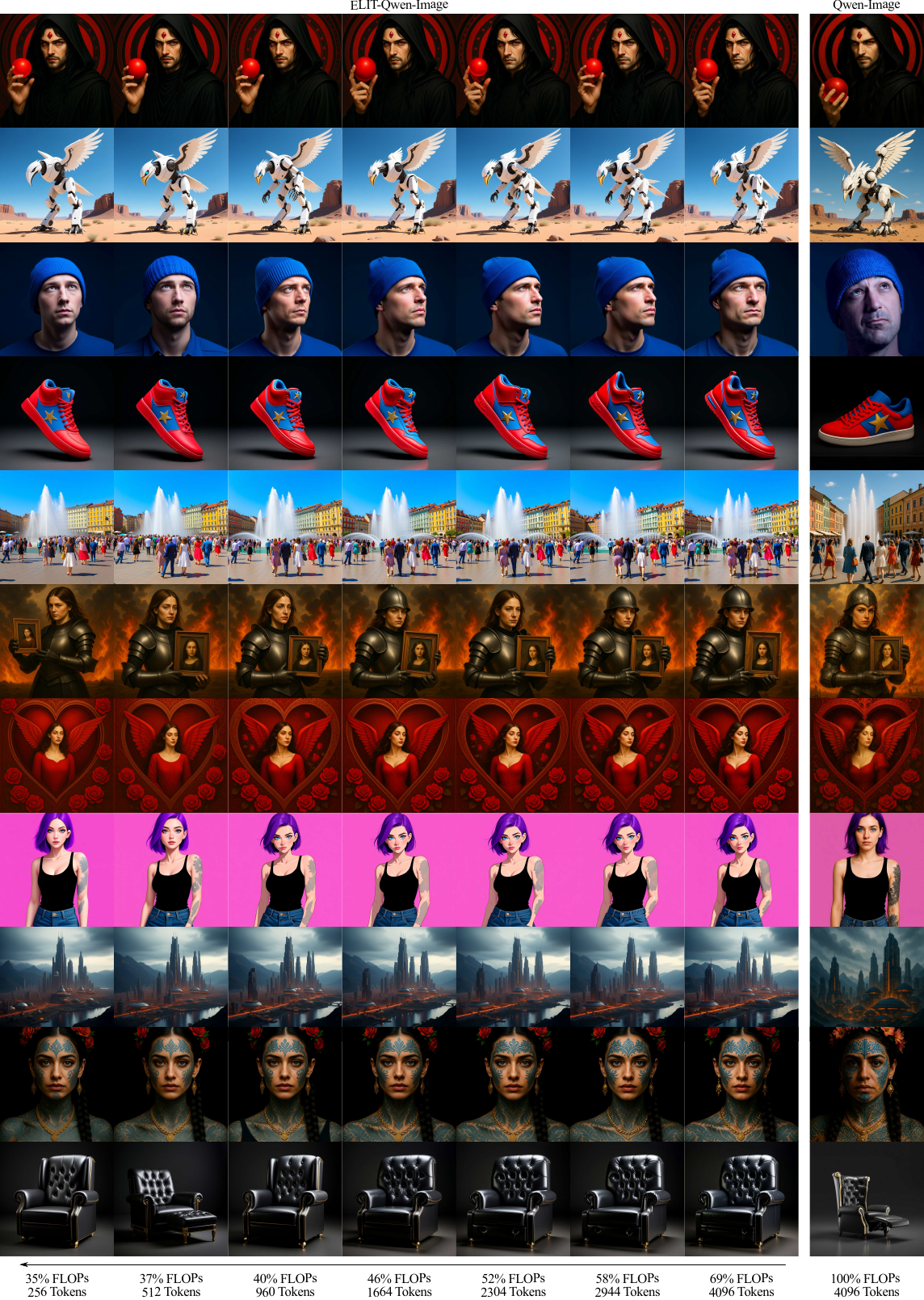}
    \caption{Qualitative results produced by \methodacronym{}-Qwen-Image for varying number of tokens in the latent interface. As the number of tokens is decreased and model FLOPs are reduced, our method can preserve structural details, while prioritizing changes in image details, preserving perceptual quality. Reported FLOPs are expressed relative to the original Qwen-Image and account for both the sampling FLOPs reductions brought by CCFG and the reduction in the number of tokens in the latent interface.} 
    \label{fig:appx_qwen_qualitatives}
\end{figure*}

\textbf{Additional Qwen-Image Results.} We provide in \Cref{fig:appx_qwen_qualitatives} additional qualitative comparison for \methodacronym{}-Qwen-Image against the original model. Thanks to CCFG, our model performs sampling with 69\% of the FLOPs with respect to Qwen-Image and is able to produce a smooth tradeoff between sample quality and model FLOPs by varying the amount of tokens in the latent interface. In the cheapest shown configuration, \methodacronym{}-Qwen-Image uses only 35\% of the FLOPs with respect to the original model. As the number of latent tokens is decreased, the model preserves structural details, prioritizing changes in the least prominent image details.

\begin{figure*}[t]
    \centering
    \includegraphics[width=0.7\linewidth]{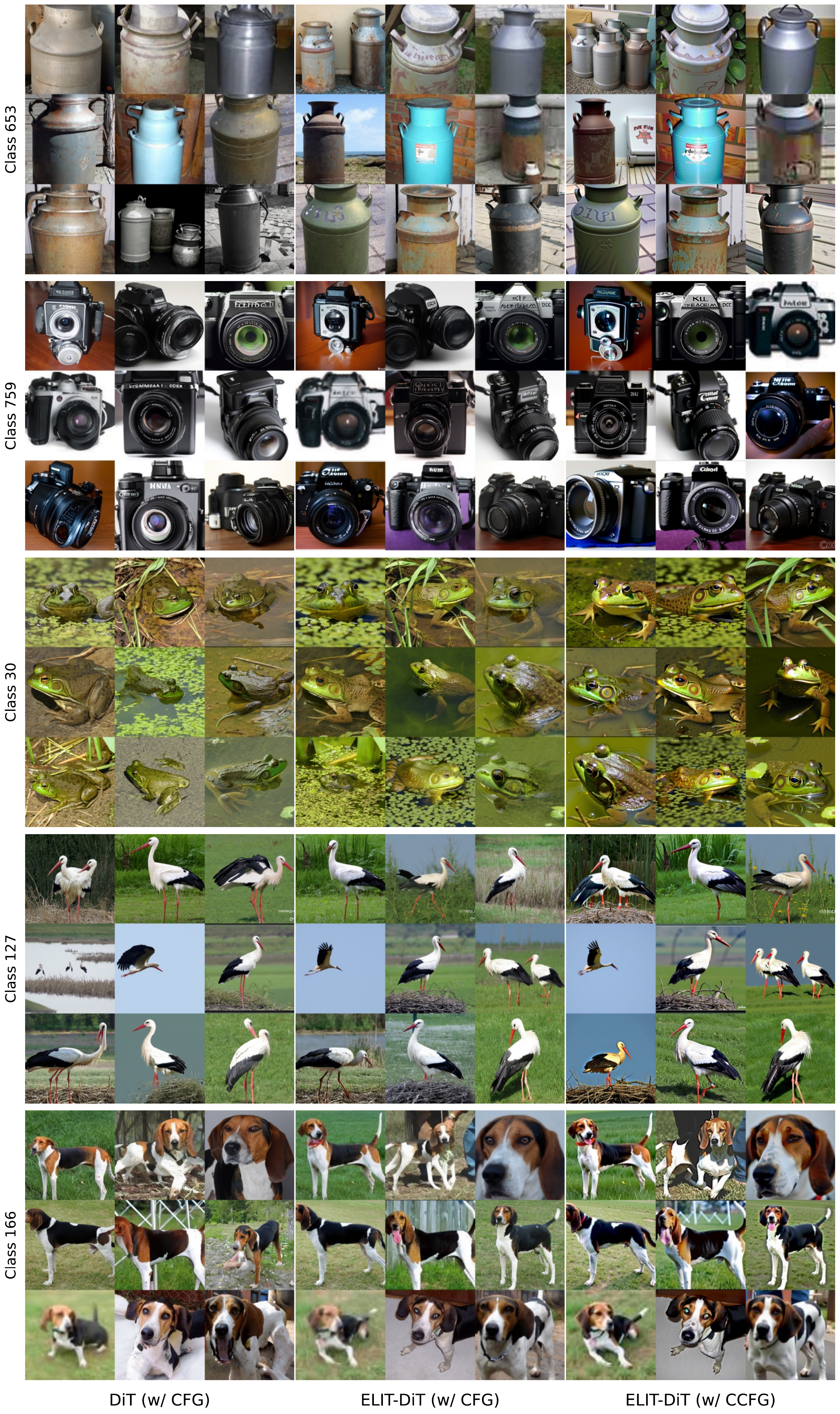}
    \caption{Uncurated Qualitative samples comparing DiT with \methodacronym{}-DiT using CFG and CCFG on ImageNet-1k 512px. Results are produced using CFG with weight 4.0 for all methods.}
    \label{fig:appx_uncurated_1}
\end{figure*}

\begin{figure*}[t]
    \centering
    \includegraphics[width=0.7\linewidth]{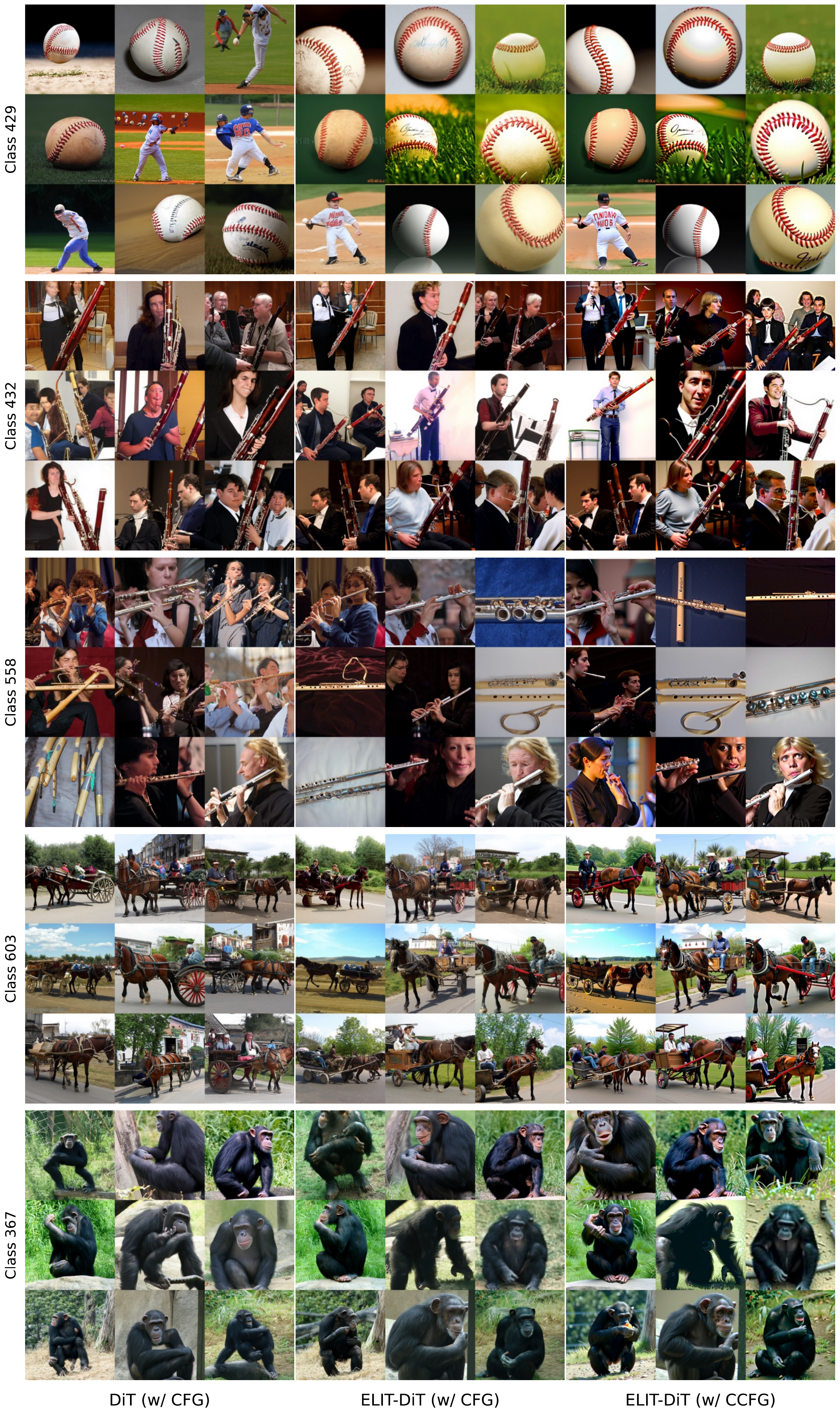}
    \caption{Uncurated Qualitative samples comparing DiT with \methodacronym{}-DiT using CFG and CCFG on ImageNet-1k 512px. Results are produced using CFG with weight 4.0 for all methods.}
    \label{fig:appx_uncurated_2}
\end{figure*}

\begin{figure*}[t]
    \centering
    \includegraphics[width=0.7\linewidth]{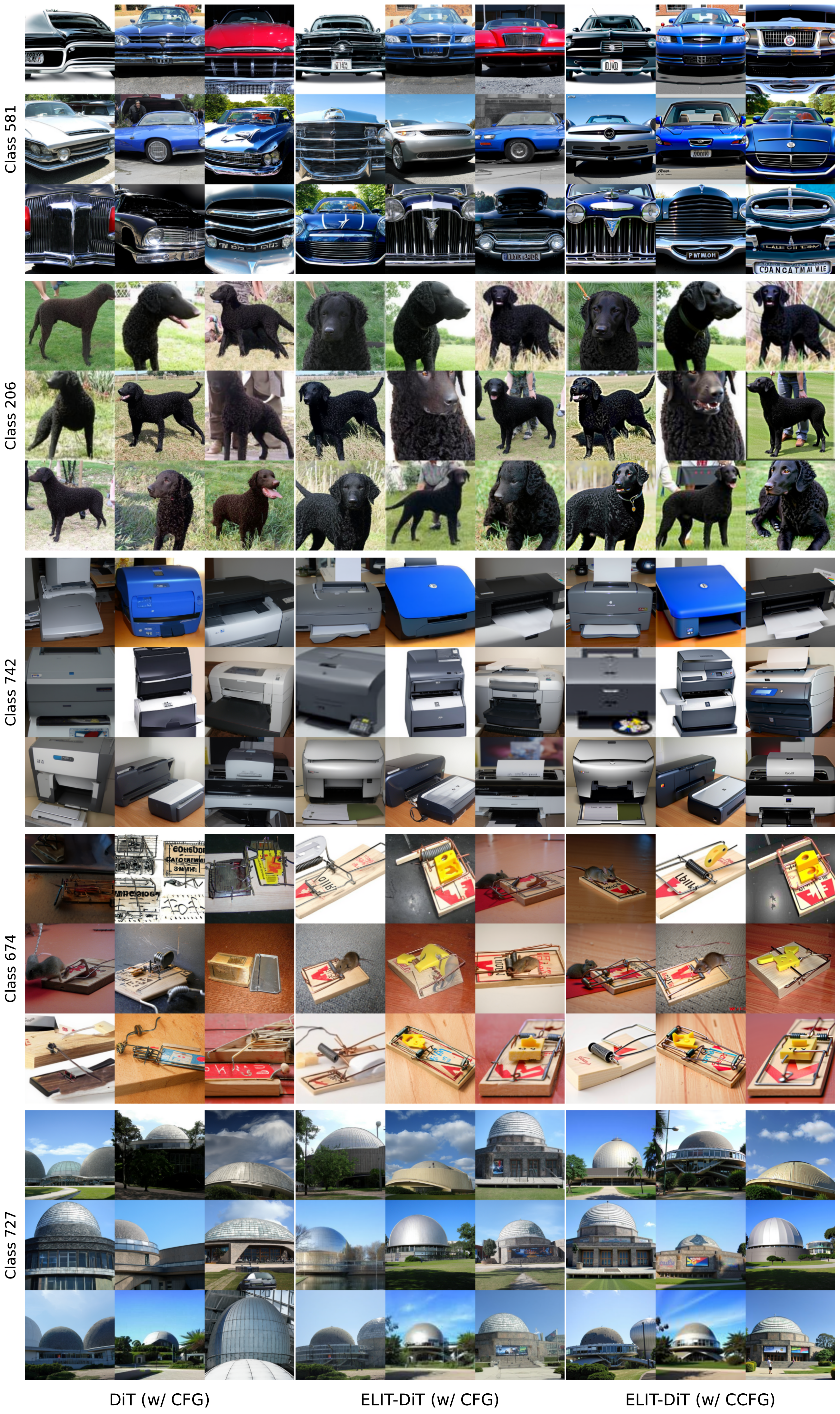}
    \caption{Uncurated Qualitative samples comparing DiT with \methodacronym{}-DiT using CFG and CCFG on ImageNet-1k 512px. Results are produced using CFG with weight 4.0 for all methods.} 
    \label{fig:appx_uncurated_3}
\end{figure*}

\textbf{Additional ImageNet-1k 512px Results.} We provide in \Cref{fig:appx_uncurated_1}, \Cref{fig:appx_uncurated_2} and \Cref{fig:appx_uncurated_3} additional qualitative comparison on ImageNet-1k 512px where we compare baseline DiT method with \methodacronym{}-DiT using CFG and CCFG. Class Ids and samples were randomly selected.

\section{Failed Experiments}

\noindent\textbf{Spatial token masking for flexible inference computation.} We explored ideas from masked diffusion transformers~\cite{zheng2024fast,gao2023maskeddiffusiontransformer,krause2025tread} as a way to obtain variable inference budget by dropping tokens in the spatial domains. We found token dropping in the spatial domain not to produce satisfactory results when applied at inference time and attribute its lower performance to the unrecoverable information loss in the spatial regions corresponding to dropped tokens. 

\noindent\textbf{Per group latent tokens count.}
We experiment with automatic per-group budget assignment, i.e. making $\latenttokcountgroupmasked$ different for each group rather than uniform across groups, with the aim of assigning more tokens to groups with more complex content, further improving compute reallocation. To achieve this, we use the loss map to supervise an additional DiT block positioned at the beginning of the DiT which predicts importance score for every group according to the loss map. Given a desired total number of tokens, we automatically distribute latent tokens to different groups, assigning more tokens to groups with higher importance score. We find this variant to increase model and implementation complexity while matching the performance of~\methodacronym{}. We hypothesize that our read operation is already tailored to read more from spatial tokens with higher loss as shown in~\Cref{fig:synthetic_experiment_combined}.

\definecolor{rowgray}{gray}{0.95}

\begin{table*}[t]
\centering
\setlength{\tabcolsep}{6pt}
\renewcommand{\arraystretch}{1.2}
\rowcolors{2}{rowgray}{white} %

\begin{tabularx}{\textwidth}{|l|X|}
\hline
\rowcolor{gray!20}
\textbf{Figure} & \textbf{Prompt} \\
\hline
\Cref{fig:teaser} & \emph{``The image portrays a woman with dark skin wearing a gold headpiece adorned with a blue jewel. Her gaze is directed towards something off-camera, giving her a focused expression. The background appears to be blurred, drawing attention to her face and headpiece.''} \\
\hline
\Cref{fig:teaser} & \emph{``The image features actor Liev Schreiber in a snowy scene from a movie or TV show. He is dressed in black tactical gear, including a vest with ``ARCTIC OCEAN'' written on it, and a helmet with goggles. The setting appears to be a bustling city street filled with people and vehicles, all covered in snow.''} \\
\hline
\Cref{fig:teaser} & \emph{``The image features a woman walking down a city street at night. She is wearing a black leather jacket, a white crop top, and a short black skirt. The street is illuminated by neon signs and streetlights, creating a vibrant atmosphere. There are other people visible in the background, but they are not the main focus of the image.''} \\
\hline
\Cref{fig:appx_qwen_qualitatives} & \emph{``The image portrays a man with long black hair and red eyes, wearing a black hooded cloak. He has a red gem on his forehead and holds a red orb-like object in his hand. The background features a circular pattern with red and black colors.''} \\
\hline
\Cref{fig:appx_qwen_qualitatives} & \emph{``The image features a large, white robot-like creature with wings standing on a desert landscape. The creature has sharp claws and appears to be looking down at something. Its body structure resembles a fusion of humanoid and bird-like characteristics. The background consists of a clear blue sky and rocky terrain.''} \\
\hline
\Cref{fig:appx_qwen_qualitatives} & \emph{``The image features a man wearing a blue knit cap, looking upwards with a serious expression. The background is dark blue, creating a contrast with the man's face and hat.''} \\
\hline
\Cref{fig:appx_qwen_qualitatives} & \emph{``The image showcases a vibrant sneaker with a red upper and blue accents. The shoe features a gold star design on the side and has red laces. The background appears to be a dark gray or black surface, providing a stark contrast to the colorful sneaker.''} \\
\hline
\Cref{fig:appx_qwen_qualitatives} & \emph{``The image captures a lively scene in a city square where people are walking around a fountain that is spraying water into the air. The square is surrounded by colorful buildings, creating a vibrant atmosphere. People are dressed in various styles of clothing, including dresses and suits, indicating a diverse crowd. Some individuals are carrying handbags, suggesting they might be tourists or shoppers. The sky above is blue''} \\
\hline
\Cref{fig:appx_qwen_qualitatives} & \emph{``The image portrays a woman dressed in full armor, holding a small picture frame with a portrait of another woman inside. The background features dramatic clouds and fire, adding intensity to the scene.''} \\
\hline
\Cref{fig:appx_qwen_qualitatives} & \emph{``The image portrays a woman inside a large, ornate heart with wings. The heart is surrounded by red roses and intricate designs, creating a fantastical and romantic atmosphere.''} \\
\hline
\Cref{fig:appx_qwen_qualitatives} & \emph{``The image portrays a woman with purple hair and tattoos on her arm. She has striking blue eyes and is wearing a black tank top and jeans. The background is a solid color, possibly pink or magenta.''} \\
\hline
\Cref{fig:appx_qwen_qualitatives} & \emph{``The image depicts a futuristic cityscape with tall buildings and domed structures illuminated by orange lights. The city is surrounded by mountains and is situated near a body of water. The sky above the city appears cloudy.''} \\
\hline
\Cref{fig:appx_qwen_qualitatives} & \emph{``The image features a woman with intricate blue tattoos on her face and neck. She has a serious expression and is adorned with gold jewelry, including earrings and a necklace. Her hair is styled in braids, and she wears a flower crown. The background is dark, which contrasts with her colorful appearance.''} \\
\hline
\Cref{fig:appx_qwen_qualitatives} & \emph{``The image features a luxurious black leather armchair with gold accents. The chair has a high backrest adorned with buttons and a footrest. It is positioned against a dark background, creating a dramatic effect.''} \\
\hline
\end{tabularx}
\caption{Prompts used to produce the showcased qualitative results for Qwen-Image~\cite{wu2025qwenimage}.}
\end{table*}

\end{document}